%% file: acl_latex.tex
\pdfoutput=1

\documentclass[11pt]{article}

 \usepackage[]{acl}

\usepackage{times}
\usepackage{latexsym}

\usepackage[T1]{fontenc}

\usepackage[utf8]{inputenc}

\usepackage{microtype}
\usepackage{array}
\usepackage{amsmath,amsthm,amsfonts,amssymb,bm,stmaryrd}
\usepackage{adjustbox}
\usepackage{algorithm}
\usepackage{algorithmic}
\usepackage{booktabs,subcaption,amsfonts,dcolumn}
\usepackage{cleveref}
\usepackage{dashbox}
\usepackage{enumitem}
\usepackage{float}
\usepackage{hyperref}
\usepackage{multirow}
\usepackage{multicol}
\usepackage{makecell}
\usepackage{CJKutf8}
\usepackage{paralist}

\crefformat{footnote}{#2\footnotemark[#1]#3}

\usepackage{placeins}
\usepackage{pifont}
\usepackage{tcolorbox}
\usepackage{tabularx}
\usepackage{url}

\usepackage{xspace}
\input{head}

\title{NSP-BERT: A Prompt-based Few-Shot Learner \\Through an Original Pre-training Task —— Next Sentence Prediction}

\author{Yi Sun\thanks{~~Equal contribution}\:\,, Yu Zheng$^{*}\text{,}$ Chao Hao, Hangping Qiu\thanks{~~Corresponding author} \\
  Army Engineering University of PLA, Nanjing, China \\
  \texttt{sunyi\_lgdx@sina.com}, \texttt{zhengyu87@outlook.com} \\
  \texttt{haochaoleo@163.com}, \texttt{qiuhp\_zy@163.com} \\
  }

\begin{document}
\maketitle
\begin{abstract}
Using prompts to utilize language models to perform various downstream tasks, also known as prompt-based learning or prompt-learning, has lately gained significant success in comparison to the pre-train and fine-tune paradigm. Nonetheless, virtually most prompt-based methods are token-level such as PET based on mask language model (MLM). In this paper, we attempt to accomplish several NLP tasks in the zero-shot and few-shot scenarios using a BERT original pre-training task abandoned by RoBERTa and other models——Next Sentence Prediction (NSP). Unlike token-level techniques, our sentence-level prompt-based method {\bf NSP-BERT} does not need to fix the length of the prompt or the position to be predicted, allowing it to handle tasks such as entity linking with ease. NSP-BERT can be applied to a variety of tasks based on its properties. We present an NSP-tuning approach with binary cross-entropy loss for single-sentence classification tasks that is competitive compared to PET and EFL. By continuing to train BERT on RoBERTa's corpus, the model's performance improved significantly, which indicates that the pre-training corpus is another important determinant of few-shot besides model size and prompt method.\footnote{Our code and pre-trained models are publicly at: \url{https://github.com/sunyilgdx/Prompts4Keras}.}
\end{abstract}

\input{sections/01_introduction}

\input{sections/02_related_work}

\input{sections/03_framework}
\input{sections/04_experiment}

\input{sections/05_conclusion}

\section{Acknowledgements}
We thank Su Jianlin for the repository bert4keras and the spirit of open source. We thank the insightful comments from the anonymous reviewers.


\bibliography{anthology,custom}
\bibliographystyle{acl_natbib}

\clearpage
\appendix
\label{sec:appendix}
\input{Appendix}

\end{document}

%% file: head.tex
\newcommand{\cls}{\texttt{[CLS]}}
\newcommand{\sep}{\texttt{[SEP]}}
\newcommand{\mask}{\texttt{[MASK]}}
\newcommand{\eos}{\texttt{[EOS]}}
\newcommand{\pad}{\texttt{[PAD]}}

\newcommand{\lab}{\texttt{[label]}}
\newcommand{\noun}{\texttt{[noun]}}
\newcommand{\pron}{\texttt{[pron]}}
\newcommand{\e}{\texttt{[e]}}
\newcommand{\isnext}{\texttt{IsNext}}
\newcommand{\notnext}{\texttt{NotNext}}

\newcommand{\entailment}{\texttt{Entailment}}

\newcommand{\notentailment}{\texttt{NotEntailment}}
\newcommand{\sent}{$\mathbf{x}$}

%% file: sections/01_introduction.tex
\section{Introduction}

\noindent GPT-2 (up to 1.5B \cite{radford2019language}) and GPT-3 (up to 175B \cite{brown2020language}) are ultra-large-scale language models with billions of parameters that have recently demonstrated outstanding performance in various NLP tasks. Compared with previous state-of-the-art fine-tuning methods, they can achieve competitive results without any or with just a limited quantity of training data. Although studies have shown that scaling up the model improves task-agnostic and few-shot performance, some studies have shown that by constructing appropriate prompts for the model, models like BERT \cite{devlin2018bert} or RoBERTa \cite{liu2019roberta} can achieve similar performance despite having a parameter count that is several orders of magnitude smaller \cite{schick2021it, schick2021exploiting, wang2021entailment}. 
\input{figures/prompts}
Since then, the area of natural language processing has seen a fresh wave of developments, including the introduction of a new paradigm known as \textbf{prompt-based learning} or \textbf{prompt-learning}, which follows the \textit{"pre-train, prompt, and predict"} \cite{liu2021pretrain} process. In zero-shot and few-shot learning, prompt-learning has achieved a lot of success. Not only does it achieve outstanding performance, prompt-learning better integrates pre-training and downstream tasks and brings NLP tasks closer to human logic and habits.

\input{figures/mlm_nsp}

The input text for the classification task, for example, ``{\sl The Italian team won the European Cup.}'', should be assigned to one of the candidate labels, such as {\sl Gaming}, {\sl Sports}, or {\sl Finance}. At this point, the template ``{\sl This is} {\tt [MASK]} {\sl news}.'' will be added to the original text, and the model will be asked to predict the missing word or span. The model's output will then be mapped to the candidate labels. We could utilize the pre-training tasks of several types of language models (LM) to predict the abovementioned templates, including but not limited to Left-to-right LM (GPT series \cite{radford2018improving, radford2019language, brown2020language}), Masked LM (BERT \cite{devlin2018bert}, RoBERTa \cite{liu2019roberta}), prefix LM (UniLM \cite{dong2019unified, bao2020unilmv2}) and Encoder-decoder LM (T5 \cite{raffel2019exploring}, BART \cite{lewis2020bart}).

Although most research on prompt-learning has been conducted, the majority of the pre-training tasks used in prompt-learning are token-level, requiring the labels to be mapped to a fixed-length token span \cite{schick2021it, schick2021exploiting, cui2021template}. On the one hand, when the number of labels grows rapidly, this necessitates a lot of human labor. On the other hand, tasks with variable-length options make Left-to-right LM (L2R LM) or masked LM (MLM) difficult to cope with. The length of each candidate entity's description, for example, varies significantly in the entity linking task.

At the same time, we observed that there is an original sentence-level pre-training object in vanilla BERT——\textbf{NSP} (\textbf{N}ext \textbf{S}entence \textbf{P}rediction), which is a binary classification task that predicts whether two sentences appear consecutively within a document or not. Many models, like RoBERTa \cite{liu2019roberta} and many others  \cite{conneau2019cross, yang2019xlnet, joshi2020spanbert}, have questioned and abandoned this task during pre-training. Nevertheless, based on the task's features and object, we believe it is appropriate to use in prompt-learning.

Unlike most prior works, we present NSP-BERT, a sentence-level prompt-learning method. The paper's main contributions can be summarized as follows:

\begin{compactitem}
\item We propose the use of NSP, a sentence-level pre-training task for prompt-learning, which can ignore the uncertain length of the label words. Our NSP-BERT has a strong zero-shot learning capacity and can be applied to a wide range of tasks, which is extremely motivating for future work.

\item We present NSP-tuning for single-sentence classification tasks.  Without abandoning the original NSP head, binary cross-entropy loss is utilized to make the zero-shot capacity of NSP-BERT continue to few-shot by building coupled positive and negative instances.

\item By using RoBERTa's corpus to continue pre-training the BERT model, although the computational cost is only about 2\% of RoBERTa, our BERT$_{\mathcal{C}_{\rm B+Mix5}}$ has been greatly improved in both zero-shot and few-shot scenarios. We believe that the effect of pre-training corpus on few-shot learning is decisive, so we suggest that all few-shot learning baselines, even if cannot use the same pre-trained model, should be based on the same pre-training corpus. In this way, a fair comparison can be made.

\end{compactitem}

\input{figures/nsp_classification}

%% file: figures/prompts.tex
\begin{figure}[t]
\centering
\includegraphics[width=1.0\columnwidth]{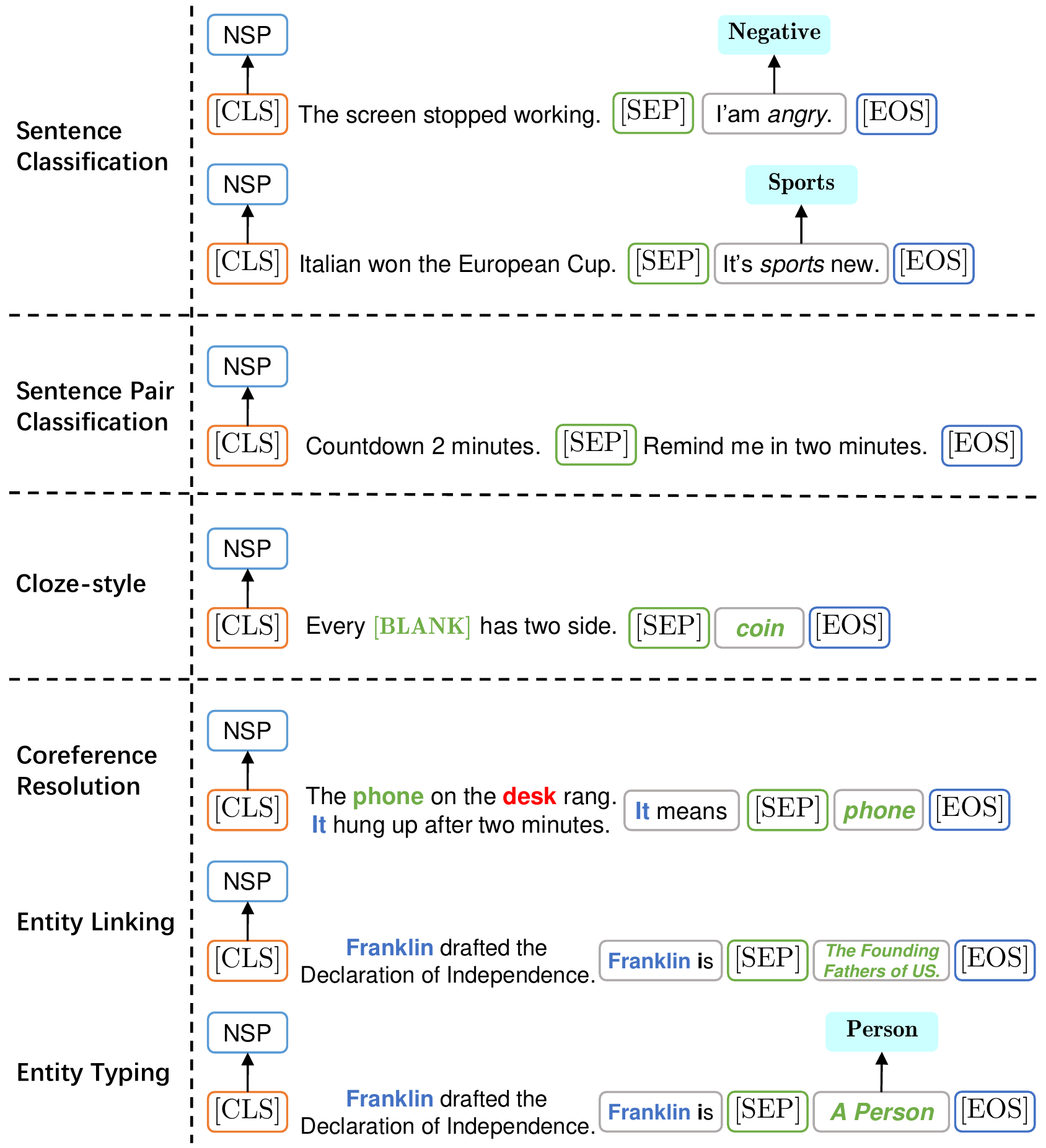} 
\caption{Prompts for various NLP tasks of NSP-BERT.}
\label{fg:prompts}
\end{figure}

%% file: figures/mlm_nsp.tex
\begin{figure*}[h]
\centering
\includegraphics[width=1.0\textwidth]{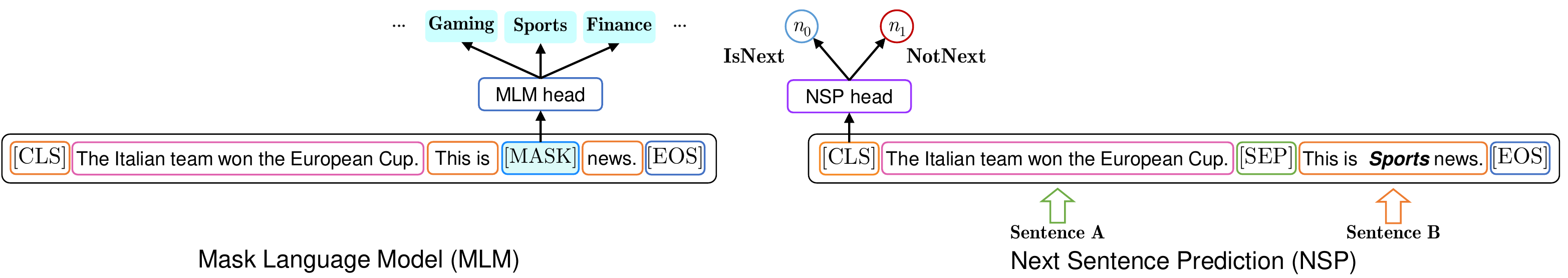} 
\caption{(Left) MLM task for token-level prompt-learning. (Right) NSP task for sentence-level prompt-learning.}
\label{fg:mlm_nsp}
\vspace{-9pt}
\end{figure*}

%% file: figures/nsp_classification.tex
\begin{figure*}[h]
\centering
\includegraphics[width=0.9\textwidth]{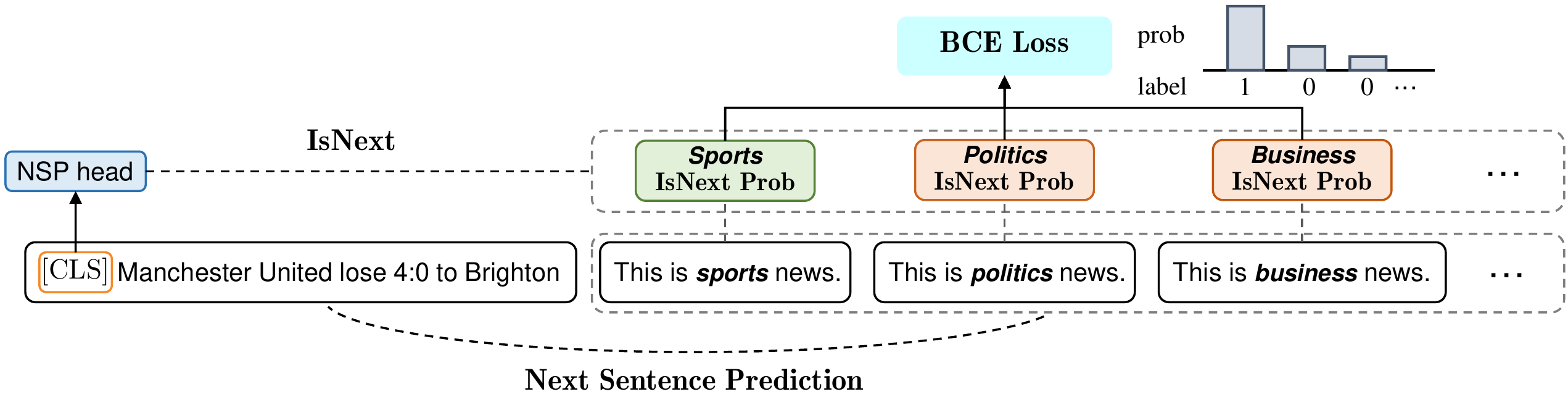} 
\caption{NSP-tuning for single-sentence classification. ``{\sl Manchester United lose 4:0 to Brighton}'' is the original input, the gold label is {\sl Sports}. The negative instances are building with wrong label {\sl Politics}, {\sl Bussiness}, etc.}
\label{fg:label_mapping}
\vspace{-9pt}
\end{figure*}

%% file: sections/02_related_work.tex
\section{Related Work}


\subsection{Token-Level and Sentence-Level}

\paragraph{Token-Level Prompt-Learning} Token-level pre-training tasks, such as MLM (Shown in the left part of Figure \ref{fg:mlm_nsp}) \cite{jiang2020how, schick2021it, schick2021exploiting} or L2R LM\cite{radford2019language, brown2020language, cui2021template}, are commonly used in token-level prompt-learning approaches. Although the expected answer may be in the form of tokens, spans, or sentences in token-level prompt-learning, the predicted answer is always generated token by token. Tokens are usually mapped to the whole vocabulary or a set of candidate words \cite{petroni2019language, cui2021template, han2021ptr,adolphs2021query, hu2021knowledgeable}. Take PET model \cite{schick2021it, schick2021exploiting} as an example, the sentiment classification input/label pair is reformulated to ``$\textbf{x}$: {\tt [CLS]} {\sl The Italian team won the European Cup. This is }{\tt [MASK]} {\sl news.} {\tt [EOS]}, $y$: {\sl Sports}''.

\paragraph{Sentence-Level Prompt-Learning} Sentence-level methods concentrate on the relationship between sentences, with the model's output usually mapped to a relationship space. As far as we know, EFL \cite{wang2021entailment} is the only sentence-level model. It reformulates NLP tasks into sentence entailment-style tasks. For example, the sentiment classification input/label pair is reformulated to ``$\mathbf{x}$: {\tt [CLS]} {\sl The Italian team won the European Cup. }{\tt [SEP]} {\sl This is Sports news.}{\tt [EOS]}, $y$: {\tt Entail}''. The output of model is {\tt Entail} or {\tt Not Entail}. The EFL model can perform well on few-shot learning but relies on labeled natural language inference (NLI) datasets like MNLI \cite{williams2018a}.

\subsection{Optimization methods}

\paragraph{Automated Prompt} Manually designed prompts are highly unstable. Sometimes it is necessary to be familiar with the particular task and language model in order to construct a high-quality prompt. As a result, several studies attempt to automatically search for and generate prompts. LM-BFF \cite{gao2021making} model use conditional likelihood to automatically select labels words, and use T5 \cite{raffel2019exploring} to generate templates. \textsc{AutoPrompt} \cite{shin2020autoprompt} uses a gradient-guided search to create prompts. Compared to the discrete prompt search methods mentioned above, P-tuning \cite{liu2021gptunderstands} employs trainable continuous prompt embeddings on GPT.

\paragraph{Training Strategy} There are many optimization methods in prompt-learning. ADAPET \cite{tam2021improving} uses more supervision by decoupling the losses for the label tokens and a label-conditioned MLM objective over the full original input. PTR \cite{han2021ptr} incorporates logic rules to compose task-specific prompts with several simple sub-prompts. \cite{zhao2021calibrate} use content-free inputs (e.g. ``N/A'') to calibrate the model's output probabilities and improved the performance of GPT-2 and GPT-3.

%% file: sections/03_framework.tex
\section{Framework of NSP-BERT}
\paragraph{Problem of MLM: Span Prediction}
As the most important pre-training task of BERT-like models, MLM has been used for prompt-learning in most previous studies, and achieved satisfactory results on GLUE \cite{wang2019glue} and other English datasets or benchmarks. In those English tasks, we can use just one token to map each label. But in some cases, we need more than one token. 
\begin{CJK*}{UTF8}{gbsn}
\begin{gather*}
    \resizebox{.75\hsize}{!}{%
    $\mathbf{x}_{input} = \cls~ {\bf x} ~\text{It was}~\mask.\eos$
    }\\
    \resizebox{1.\hsize}{!}{%
    $\mathbf{x}_{input} = \cls~ {\bf x} ~\text{这是}~\mask\mask\text{新闻}.\eos$
    }
\end{gather*}
\end{CJK*}
\begin{CJK*}{UTF8}{gbsn}
As shown in the above example, in the first English sample, ${\bf x}$ is the original sentence, we can use just one \mask token to predict the label word ``Sports'' in a classification task. But in the second Chinese sample, we need \mask\mask to map the label word ``体育'' (which has the same meaning with ``Sports''), and use their probability product to represent the probability of the label (detailed description is in the Appendix \ref{sec:probs} ). As the number of \mask increases, it becomes difficult for the MLM to predict correctly. At the same time, it is impossible to compare the probability of label mapping words (spans or sentences) with different number of \mask tokens, entity linking is one of the scenarios. Therefore, especially in the Chinese task, there is a obvious gap between the pre-training and the downstream task.
\end{CJK*}
\subsection{Next Sentence Prediction}
The next sentence prediction is one of the two basic pre-training tasks (the other is MLM) of the vanilla BERT model \cite{devlin2018bert} (Shown in the right part of Figure \ref{fg:mlm_nsp}). This task inputs two sentences {\tt A} and {\tt B} into BERT at the same time to predict whether sentence {\tt B} comes after sentence {\tt A} in the same document. During specific training, for $ 50\% $ of the time, {\tt B} is the actual next sentence that follows {\tt A} ({\tt IsNext}), and for the other $ 50\% $ of the time, we use a random sentence from the corpus ({\tt NotNext}).

\begin{equation*}
\setlength{\abovedisplayskip}{1pt}
\setlength{\belowdisplayskip}{1pt}
\mathbf{x}_{input}=\cls\mathbf{x}_i^{(1)}\sep\mathbf{x}_i^{(2)}.\eos
\label{eq:nsp_input}
\end{equation*}

Let ${\mathcal{M}}$ denote the model trained on a large-scale corpus. This model is trained on both MLM task and NSP task at the same time. $\mathbf{x}_i^{(1)}$ and $\mathbf{x}_i^{(2)}$ denote sentence {\tt A} and sentence {\tt B}, respectively. The model's input is $\mathbf{x}_{input}$, and $q_{\mathcal{M}}$ denotes the output probability of model's NSP head.  $\mathbf{s}=\mathbf{W}_{\mathrm{nsp}}(\tanh{(\mathbf{W}\mathbf{h}_{[ \mathrm{CLS} ]}+\mathbf{b})})$ \footnote{Devlin et al.\shortcite{devlin2018bert} use an additional nonlinear layer to pool the hidden vector of \cls for NSP task, but not mentioned in their paper.}, where $\mathbf{h}_{[ \mathrm{CLS} ]}$ is the hidden vector of 
\cls and $\mathbf{W}_{\mathrm{nsp}}$ is a matrix learned by NSP task, $\mathbf{W}_{\mathrm{nsp}}\in \mathbb{R} ^{2\times H}$. The loss function of NSP task $\mathcal{L} _{\mathrm{NSP}} = -\log q_{\mathcal{M}}( n|\mathbf{x})$, where $n\in \{ \isnext, \notnext \}$.
\begin{equation}
q_{\mathcal{M}}( n_k|\mathbf{x}_i ) =\frac{\exp s( n_k|\mathbf{x}_{i}^{( 1 )},\mathbf{x}_{i}^{( 2 )} )}{\sum\limits_{n}{\exp s( n|\mathbf{x}_{i}^{( 1 )},\mathbf{x}_{i}^{( 2 )} )}}
\label{eq:nsp_output}
\end{equation}

\subsection{Prompts in NSP-BERT}
NSP-BERT, like other prompt-based learning methods, requires the construction of appropriate templates for various tasks. In order to make the model have better zero-shot performance and better few-shot initialization, the template's building form must closely match the original NSP task. In this section, we'll show how to construct templates for different tasks (also shown in Figure \ref{fg:prompts}).

In order to apply NSP to zero or few-shot learning, we treat most tasks as multiple-choice tasks. Same as the right side in Figure \ref{fg:mlm_nsp}, an NSP-BERT's input can be expressed as:
\begin{equation*}
\resizebox{.85\hsize}{!}{%
$\mathbf{x}_{input}=\cls \mathbf{x}_i \sep p_{i}^{(j)}\eos$.
}
\label{eq6}
\end{equation*}
We define the template $\mathcal{T}$ as a combination of input $\mathbf{x}_i$ and the prompts, $\mathcal{T}(\mathbf{x})=$ \cls $\mathbf{x}$ \sep {\sl This is} ... {\sl news}.\eos. Unlike prompt-tuning based on MLM \cite{schick2021exploiting,gao2021making} which requires mapping labels to vocabularies, for our NPS-BERT, labels can be mapped to words or phrases of arbitrary length in ``...''. To map labels to the prompts, we define a verbalizer as a mapping $f: \mathcal{Y} \mapsto \mathcal{P}$. The label $y_{i}^{(j)}$ can be mapped to prompt $p_{i}^{(j)} \in \mathcal{P} $.

In single-sentence classification tasks, all samples share the same label space $\mathcal{Y}$, where $| \mathcal{Y}|$ is the number of classes. For label of the $j$th class $y^{(j)} \in \mathcal{Y} $ can be mapped to prompt $p^{(j)}$. For those tasks where each sample corresponds to different labels, such as cloze-style task, word sense disambiguation, entity linking, we define the label space corresponding to the $i$th sample as $\mathcal{Y}_i$, and $y_{i}^{(j)} \in \mathcal{Y}_{i} $.
\input{figures/two_stage}

In tasks such as entity linking, there are more than one entity in the sentence, in order to identify target entity words, we recommend using {\bf two-stage prompt} (as shown in Figure \ref{fg:two_stage}) to indicate the target word using natural language descriptions:
\label{sec:wsd}
\begin{compactitem}
    \item {\bf Stage 1}: Prompt the target word at the end of sentence {\tt A}. This stage's purpose is to provide enough context for the target word.
    \item {\bf Stage 2}: Prompt the description of the candidate word sense in sentence {\tt B}. 
\end{compactitem}
Let $p_{i,1}^{(j)}$ and $p_{i,2}^{(j)}$ denote the first and the second part of the prompt. The model's input is:
\begin{equation*}
\mathbf{x}_{input}=\cls \mathbf{x}_i,p_{i,1}^{(j)} \sep p_{i,2}^{(j)}\eos.
\end{equation*}

For sentence-pair tasks such as text entailment and text matching, since the NSP task is in the form of sentence pairs we still use the same input as the original NSP task. 

\subsection{Answer Mapping} Because not all datasets can provide contrastive candidate answers (sentiments, topics, idioms, or entities), we propose two answer mapping methods, {\bf candidates-contrast} answer mapping and {\bf samples-contrast} answer mapping, for different situations.

\paragraph{Candidates-Contrast} For datasets with multiple candidates, such as candidate sentiments, candidate topics, candidate idioms and candidate entities. For the above datasets, there is a template $p_{i}^{(j)}$ (or $p_{i}$) corresponding to the label $y_{i}^{(j)}$ (or $y_{i}$), we choose the {\tt IsNext} probability as the output of each candidate answer. The logit of label $y_{i}^{(j)}$ (the value ranges from 0 to 1, but is not an actual probability) is:
\begin{equation}
\setlength{\abovedisplayskip}{3pt}
\setlength{\belowdisplayskip}{3pt}
\displaystyle
q( y^{ ( j )}_{i}|\mathbf{x}_i ) \propto { q_{\mathcal{M}}(n= \isnext |\mathbf{x}_i,p^{( j )}_{i})}
\end{equation}
In the prediction stage, we take the highest probability output by ${\mathcal{M}}$ among the candidates as the final output answer where the condition is {\tt IsNext}:
\begin{equation}
\setlength{\abovedisplayskip}{3pt}
\setlength{\belowdisplayskip}{0pt}
\resizebox{.85\hsize}{!}{%
$\begin{aligned}
    \hat{y}_{i}
    &=\arg\max\limits_{j}\,q(y_{i}^{( j )}| \mathbf{x}_i ) \\
    &=\arg\max\limits_{j}\,q_{\mathcal{M}}( n=\isnext|\mathbf{x}_i,p_{i}^{( j )} ) 
\end{aligned}$}
\end{equation}
\input{tables/main_results_cls}
\paragraph{Samples-Contrast} For sentence-pair tasks, the \isnext~output probabilities of most samples are close to $1$ (see details in Appendix \ref{sec:prob_sentence_pair}), which makes it difficult to judge the relationship between two sentences through a single sample. So we propose the samples-contrast answer mapping method (Figure \ref{fg:label_mapping}), to determine the label of a individual sample by contrast the probability of \isnext~between samples. To put it simply, by {\bf rank}ing\footnote{Sort samples in ascending or descending order according to \isnext~probability.} in ascending order, the samples with a relatively higher \isnext~probability are {\bf divide}d\footnote{Divide the dataset (or sample batch) into subsets according to the proportion of each label in development set.} into labels with a higher degree of matching, such as \entailment. On the contrary, samples with lower \isnext~probability will be divided to labels such as \notentailment. This procedure is summarized in Algorithm \ref{alg:sample_contrast}\footnote{This method is currently only suitable for sentence-pair tasks, and can only be applied in zero-shot scenarios.}. 

Considering the fairness of the comparative experiment, we consider two preconditions. One is that a complete development set and a test set can be obtained at the same time; the other is that only the development set can be obtained, and the test samples must be predicted one by one or batch by batch during testing. In our experiment, we use the development set to determine the thresholds of probability, and use these thresholds to predict the test set.

\input{algorithm/sample_contrast}

\subsection{NSP-tuning}
Since we treat tasks with candidates as multiple-choice problems, when we need to perform few-shot learning, we need to choose some methods to continue the initialization advantages of NSP-BERT in zero-shot. We name this method NSP-tuning used on few-shot single-sentence classification tasks, as shown in Figure \ref{fg:label_mapping}.

\paragraph{Building Instances}
Taking the single-sentence classification as an example, for the $i$th sample, we take it's gold label $y_{i}^{+}$ as a positive instance $(\mathcal{T}(\mathbf{x}_{i},y_{i}^{+}),1)$ , while taking the rest of the labels in $\mathcal{Y}$ as negative instances $\{( \mathcal{T}(\mathbf{x}_{i},y_{i}^{-}),0 )\}^{|\mathcal{Y}|-1}_{y_{i}^{-}\ne y_{i}^{+},y_{i}^{-}\in \mathcal{Y}}$ and $\{0,1\}$ represent the labels of the binary classification. Both the positive instance and negative instances of  the same sample, a total of $|\mathcal{Y}|$, will be coupled and input to the model in a same batch.

\paragraph{Loss function} 
Since the output probability of {\texttt{IsNext}} has been already normalized to $[0,1]$ by softmax after a nonlinear layer during pre-training, if we want to do NSP-tuning without changing the structure of the pre-training model, we need to choose the {\bf binary cross-entropy loss} as the loss function. Of course, we can re-initialize the output of $\mathcal{M}$ to implement a multiple-choice method with linear layer$+$softmax cross-entropy loss same as \cite{Radford2018ImprovingLU}, but we think this is not conducive to preserving the zero-shot advantage of NSP to few-shot.

%% file: figures/two_stage.tex
\begin{figure}[h]
\centering
\includegraphics[width=1.0\columnwidth]{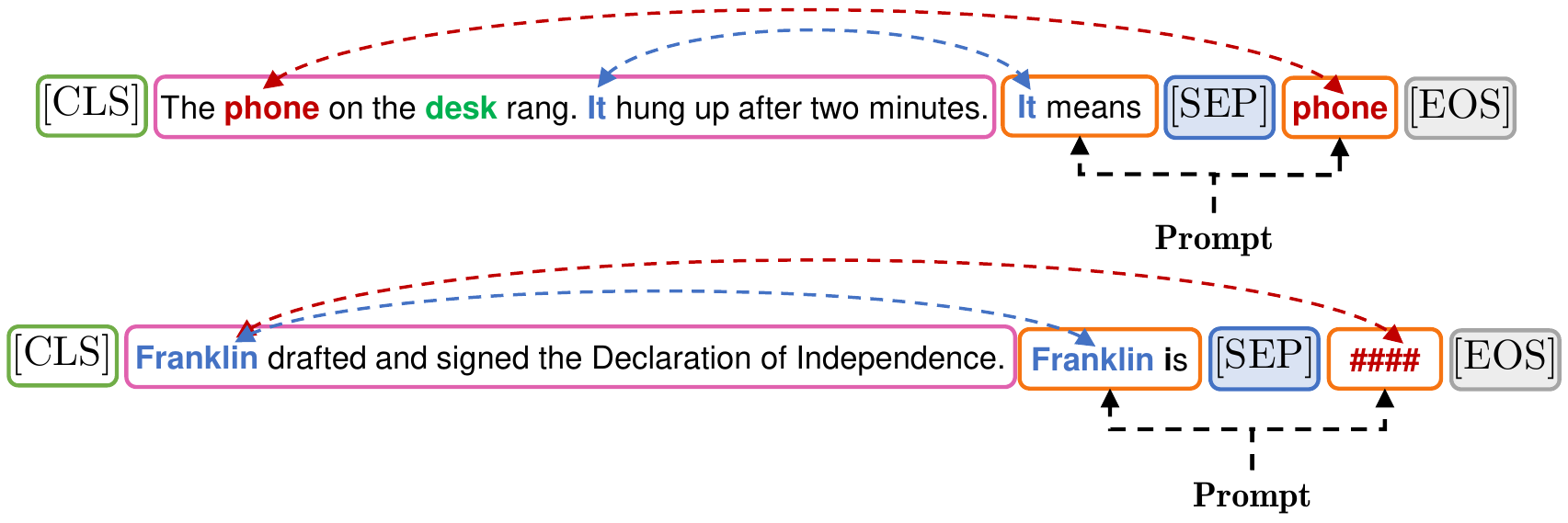} 
\caption{Two-stage prompt, examples in coreference resolution and entity linking/typing tasks.}
\label{fg:two_stage}
\vspace{-6pt}
\end{figure}

%% file: tables/main_results_cls.tex
\begin{table*}[h]
    \centering
    \resizebox{1.0\textwidth}{!}{
    \begin{tabular}{llccccccccccc}
        \toprule
         & &\multicolumn{7}{c}{\bf English Tasks} &\multicolumn{4}{c}{\bf Chinese Tasks} \\
        \cmidrule(lr){3-9}\cmidrule(lr){10-13}
        
        & & {\bf SST-2} & {\bf MR} & {\bf CR} & {\bf MPQA} & {\bf Subj} &  {\bf Yahoo!} & {\bf AGNews} & {\bf EPR.}  & {\bf TNEWS(K)}  & {\bf CSLDCP} &  {\bf IFLY.}\\
       \midrule
       
        
        \multicolumn{1}{l}{\multirow{2}{*}{Full}} & Majority & {\it 50.9} & {\it 50.0} & {\it 50.0} & {\it 50.0} & {\it 50.0} & {\it 10.0} & {\it 25.0} & {\it 50.0} & {\it 6.7} & {\it 1.5} & {\it 0.8} \\
        
        & Fine-Tuning & {\it 93.6} & {\it89.0} & {\it 89.3} & {\it 89.3} & {\it 97.0} & {\it 76.5} & {\it 94.7} & {\it 90.0}$^\dagger$ & {\it 71.0}$^\dagger$ & {\it 68.0}$^\dagger$ & {\it 66.0}$^\dagger$ \\
     
        \midrule
        
        
        \multicolumn{1}{l}{\multirow{2}{*}{Zero}} & PET & {67.6} & {65.3} & \underline{\bf 61.2} & \underline{\bf 63.9} & \underline{\bf 61.0} & {25.6} & {54.5} & {60.7} & {28.0 / 35.6} & {22.4}  & {34.8} \\
         
        & NSP-BERT & \underline{\bf 75.6} & \underline{\bf 74.4} &	{59.4} & {59.9} & {53.9}  & \underline{\bf 47.0} & \underline{\bf 77.5} & \underline{\bf 86.9} & \underline{\bf 51.9} / \underline{\bf57.0} & \underline{\bf 47.6} & \underline{\bf 41.6} \\
        
        \midrule
        
        
        \multicolumn{1}{l}{\multirow{5}{*}{Few}} & Fine-tuning & {77.9{\small ±5.9}} & {68.0{\small ±9.4}} & {79.1{\small ±8.9}} & {65.2{\small ±6.3}} & \underline{\bf 89.7{\small ±1.1}} & {61.8{\small ±1.5}} & {82.4{\small ±1.2}} & {78.7{\small ±5.8}} & {51.1{\small ±1.1} / 58.0{\small ±1.4}} & {51.7{\small ±2.1}} & {45.1{\small ±2.2}} \\
        
        & PET & {86.0{\small ±1.6}} & {80.0{\small ±1.6}} &	\underline{\bf 88.9{\small ±0.6}} & {83.3{\small ±2.4}} & {86.2{\small ±1.5}}  &{64.3{\small ±1.3}} & {84.2{\small ±0.8}} & {82.5{\small ±2.0}} & {54.7{\small ±1.1} / 61.2{\small ±0.9}} &	{52.6{\small ±1.2}} & {45.9{\small ±2.1}} \\
         
        & EFL w/\;\;\,PT & \underline{\bf 86.9{\small ±1.8}} & \underline{\bf 80.6{\small ±1.2}} & {88.1{\small ±0.9}} & \underline{\bf 86.1{\small ±0.7}} & {86.0{\small ±3.3}}  &{63.0{\small ±1.2}} & {83.8{\small ±1.3}} & {84.8{\small ±1.6}} & {53.2{\small ±1.5} / {59.2\small ±1.6}} & {52.0{\small ±1.6}} & {47.9{\small ±1.5}} \\
        
        & EFL w/o PT & {81.2{\small ±5.1}} & {76.1{\small ±9.1}} & {79.2{\small ±4.0}} & {79.1{\small ±1.6}} & {75.1{\small ±9.4}}  &{60.8{\small ±4.2}} & {84.6{\small ±0.7}} & {84.6{\small ±2.1}} & {54.7{\small ±1.3} / 60.3{\small ±1.7}} & {53.8{\small ±0.9}} & {49.5{\small ±1.2}} \\
         
        & NSP-BERT & {86.8{\small ±1.3}} & {80.5{\small ±1.5}} & {86.0{\small ±2.2}} & {83.9{\small ±1.1}} & {86.4{\small ±1.8}}  & \underline{\bf 64.5{\small ±0.5}} & \underline{\bf 85.9{\small ±0.8}} & \underline{\bf 87.7{\small ±0.7}} & \underline{\bf 55.7{\small ±1.0}} / \underline{\bf 61.6{\small ±0.9}} & \underline{\bf 55.0{\small ±1.5}} & \underline{\bf 49.5{\small ±1.1}} \\
        \bottomrule
    \end{tabular}
    }
    \caption{Main zero-shot and few-shot learning results on single-sentence classification tasks. In addition to the accuracy, we also report the standard deviation for few-shot learning. For English tasks, we use vanilla BERT-{\sc Large}. For Chinese tasks, we use UER's Chinese BERT-{\sc Base}. Full: full training; Zero: zero-shot; Few: few-shot; $\dagger$: human performance; Majority: majority class; EFL w/ PT: few-shot tuning of EFL with pre-training on MNLI; EFL w/o PT: few-shot tuning of without pre-training on MNLI; TNEWS(K): use the keyword (K) field or not.}
    \label{tb:main_result}
\end{table*}

%% file: algorithm/sample_contrast.tex
\begin{algorithm}[tb]
\caption{Samples-Contrast Answer Mapping}
\label{alg:sample_contrast}
\textbf{Input}: Test set $\mathcal{D} =\{\mathbf{x}_{i}\} _{i=1}^{N}$, where $\mathbf{x}_{i}=(\mathbf{x}_{i}^{( 1 )}, \mathbf{x}_{i}^{( 2 )})$, Oder $o \in$ \{``ascending'', ``descending''\}, distribution of labels $d$, batch size $bs$.\\
\textbf{Output}: $\{\mathbf{x}_{i}, \hat{y}_i\} _{i=1}^{N}$
\begin{algorithmic}[1] 
\FOR{$i=1,...,N$}
\STATE $q_{i} \leftarrow q_{\mathcal{M}}(n=\mathrm{IsNext}|\mathbf{x}_{i}^{(1)},\mathbf{x}_{i}^{(2)})$
\ENDFOR
\STATE $\{\mathcal{B}_{j}\}_{j=1}^{\lceil \frac{N}{bs} \rceil} \leftarrow $ {\rm \bf divide} ($\mathcal{D}, bs$) 
\FOR{$j=1,...,\lceil \frac{N}{bs} \rceil$}
\STATE $\mathcal{B}_{j}’=\{\mathbf{x}_{r(1)},...,\mathbf{x}_{r(bs)}\} \leftarrow $ {\rm \bf rank}($\mathcal{B}_{j}$, $q_{i}, o$)
\STATE $\{B_{m}\}_{m=1}^{M} \leftarrow $ {\rm \bf divide} ($\mathcal{B}_{j}’, d$)
\FOR{$i=1,...,bs$}
\STATE $\hat{y}_i \leftarrow m$ {\rm \bf where} $\mathbf{x}_{i} \in B_{m}$
\ENDFOR
\ENDFOR
\end{algorithmic}
\end{algorithm}

%% file: sections/04_experiment.tex
\section{Experiment}

\subsection{Tasks and Datasets}
\paragraph{English Datasets} For English tasks, following \cite{gao2021making,hu2021knowledgeable,liang2022contrastive}, we choose 7 single-sentence and 5 sentence-pair English tasks. See details in Appendix \ref{sec:appendix_datasets}. 

\paragraph{Chinese Datasets} For Chinese tasks, we choose FewCLUE \cite{xu2021fewclue}, a Chinese Few-shot Learning Evaluation Benchmark, which contains 9 NLU tasks in Chinese, with 4 single-sentence tasks, 3 sentence-pair tasks and 2 reading comprehension tasks. Additionally, we select the entity linking dataset DuEL2.0\footnote{https://aistudio.baidu.com/aistudio/competition/detail/83} to verify the word sense disambiguation ability. And we divide DuEL2.0 into two parts: DuEL2.0-L (entity linking) and DuEL2.0-T (entity typing).

\input{tables/corpus_matters}

\subsection{Baselines}

\paragraph{Fine-Tuning} Standard fine-tuning of the pre-trained language model on the FewCLUE training set. The models are fine-tuned with cross entropy loss and using the BERT-style model's hidden vector of {\tt [CLS]} $\mathbf{h}_{[ \mathrm{CLS} ]}$ with a classification layer ${\rm softmax}(\mathbf{W}\mathbf{h}_{[ \mathrm{CLS} ]})$, where $\mathbf{W}\in \mathbb{R} ^{|\mathcal{Y}|\times H}$, $|\mathcal{Y}|$ is the number of labels.

\paragraph{Prompt-based methods} Since our method is a brand-new basic prompt-learning method, our main purpose is to demonstrate its effectiveness compared to MLM-like methods, and we think it is not necessary to compare with more complex methods such as continuous prompt or automatic prompt methods. Therefore we choose token-level model PET \cite{schick2021it, schick2021exploiting} based on MLM and sentence-level model EFL\footnote{We use MNLI\cite{williams2018broad_mnli} and OCNLI\cite{ocnli} to pre-train EFL.} \cite{wang2021entailment} based on entailment as two baselines. 

\subsection{Experiment Settings}

\paragraph{Evaluation Protocol}

For few-shot learning, we follow the evaluation protocol adopted in \cite{gao2021making,liang2022contrastive} and assume $K$ samples per class for training set. For English tasks the $K$ of training set is set to 16, and the size of the development set is 10 times the size of the training set. The number $K$ of FewCLUE has been set to 8 or 16 according to Xu et al. \shortcite{xu2021fewclue}. For each experiment, we run 5 experiments with 5 different training and development set (split by 5 fixed random seed) and report the average results and standard deviations.

\paragraph{Language Models}
In order to conduct comparative experiments fairly, for our main experiments, we use the same pre-trained language model for the same dataset. For English tasks, we adopt the vanilla English BERT-{\sc Large}\footnote{https://github.com/google-research/bert \label{fn:google_bert}}. For Chinese tasks, we adopt the Chinese BERT-{\sc Base}\footnote{https://github.com/dbiir/UER-py \label{fn:uer}} trained by UER using MLM and NSP \cite{zhao2019uer}.

\paragraph{Hyper-parameters}
For few-shot learning, we train 10 epochs on all the datasets. We set learning rate as 2e-5 for English tasks, and 1e-5 for Chinese tasks. The batch size is 8. All baselines use the same hyper-parameters described above.

\subsection{Main Results}
The Table \ref{tb:main_result} reports the main results on 7 English and 4 Chinese single-sentence classification tasks. Since we use the same pre-trained language model for all methods, this experiment is fair enough. It is clear that our NSP-BERT offers distinct advantages in zero-shot scenario, particularly for multi-topic classification tasks such as Yahoo!, AGNews, and all Chinese datasets. In few-shot scenario. its performance is comparable to the MLM-based PET \cite{schick2021exploiting} on the most datasets. Compared with EFL \cite{wang2021entailment} without pre-training on the NLI dataset, NSP-BERT is much better. Our NSP-BERT has the fastest convergence speed  based on convergence curves, as shown in Figure \ref{fg:fitting_process}. NSP-BERT usually achieves the best performance during the first few epochs.

\input{figures/fitting_process}
\input{tables/ablations.tex}
\paragraph{Ablation studies on NSP-tuning}
It can be seen from Table \ref{tb:ablations} that coupling positive and negative samples $+$ BCE loss function is the most effective and robust way of NSP-tuning. Other modifications in the table will degrade the performance of the model and make the results unstable. We believe this is due to the special output of the NSP Head, and re-initialization will lose the knowledge gained during pre-training.

\paragraph{Impact of Pre-training Corpus} 
Compared with the RoBERTa model, the original BERT model has a large gap in the pre-training corpus. BERT is only pre-trained on Wikipedia and BookCorpus\cite{zhu2015aligning}, and the size is about 16GB, while RoBERTa additionally uses CC-News\footnote{https://commoncrawl.org/2016/10/news-dataset-available/}, OpenWebText \cite{Gokaslan2019OpenWeb} and Stories\cite{trinh2018simple} corpus, which is 145GB more.
\input{figures/histogram}
We use the above 5 corpora\footnote{Since there is no public Stories corpus, we refer to the construction method of \cite{trinh2018simple} and build it on the basis of CC-100 \cite{conneau-etal-2020-unsupervised}.} to pre-train the vanilla BERT model incrementally. Due to the limited computing power, our total training steps are about 30$\%$ of the BERT model and 2$\%$ of the RoBERTa model. As shown in Table \ref{tb:corpus_matters}, although it has not yet reached the level of RoBERTa, our BERT model (BERT$_{\mathcal{C}_{\rm B+Mix5}}$) has greatly improved the performance of zero-shot and few-shot learning, and this improvement even exceeds the changes brought by the prompt method.

\paragraph{Impact of Model Size}
Only under the premise of fixing the same pre-training corpus, we can verify the effect of model size on NSP-BERT. We carefully selected 4 sizes of UER’s BERT (tiny, small, base and large) trained on same corpus for validation on two datasets, EPRSTMT and TNEWS. Figure \ref{fg:model_size} shows the impact of different sizes of models on NSP-BERT and PET, it can be seen that our method is still very competitive on small models\footnote{PET fails to fit on tiny and small models for no reason.}.
\input{figures/model_size}

\subsection{Applications of NSP-BERT}
We validate applications of NSP-BERT on the tasks shown in Table \ref{tb:applications}, including NLI (OCNLI, MNLI, SNLI, QNLI and RTE), text matching (BUSTM), keyword recognition (CSL), Chinese idiom cloze test (ChID), and coreference resolution (CLUEWSC). In these tasks, the zero-shot learning prediction ability of NSP-BERT is demonstrated with the help of the sample-contrast method. From Figure \ref{fg:histogram}, we can see that even a small contrast batch size can help the sentence-pair tasks, and as the batch size increases, this improvement becomes more obvious and tends to be stable.

Our NSP-BERT can be applied to the task of entity typing, and can even handle entity linking task. The difficulty of entity linking for MLM-based model such as PET is that the description of the entity is of variable length. In these tasks with more than one target words or entity, the effect of two-stage prompt is obvious, see Table \ref{tb:two_stage_prompt}.

\input{tables/applications}
\input{tables/two_stage_prompt}

%% file: tables/corpus_matters.tex
\begin{table*}[h]
    \centering
    \resizebox{0.85\textwidth}{!}{
    \begin{tabular}{llllccccccc}
        \toprule
         & & \multicolumn{1}{l}{\multirow{2}{*}{\bf Model}} & \multicolumn{1}{l}{\multirow{2}{*}{\bf Corpus}} &\multicolumn{7}{c}{\bf English Tasks} \\
        
        \cmidrule(lr){5-11}
        & & & & {\bf SST-2} & {\bf MR} & {\bf CR} & {\bf MPQA} & {\bf Subj} &  {\bf Yahoo!} & {\bf AGNews} \\
        \midrule
        
        
        \multicolumn{1}{l}{\multirow{6}{*}{Zero}} & \multicolumn{1}{l}{\multirow{4}{*}{PET}} & \multicolumn{1}{l}{\multirow{2}{*}{RoBERTa}} & $\mathcal{C}_{\rm B}$  & {81.2} & {75.6} &	{76.6} & {63.3} & \underline{\bf 63.6} & {18.7} & {47.8} \\
        
        & & &  $\mathcal{C}_{\rm R}$ & \underline{\bf 83.6} & \underline{\bf 80.8} & \underline{\bf 79.5} & {\bf 67.6} & {53.6} & {\bf 25.6} & {\bf 54.5} \\
        
        \cmidrule(lr){3-11}
        & & \multicolumn{1}{l}{\multirow{2}{*}{BERT}} & $\mathcal{C}_{\rm B}$ & {67.6} & {65.3} &	{61.2} & {63.9} & {\bf 61.0} & {25.6} & {\bf 54.5} \\
        
        & & & $\mathcal{C}_{\rm B+Mix5}$ & {\bf 75.0} & {\bf 70.1} & {\bf 67.4} & {\bf  64.2} & {55.3} & {\bf 28.5} & {38.4} \\
        
         \cmidrule(lr){2-11}
        
        & \multicolumn{1}{l}{\multirow{2}{*}{NSP-BERT}} & \multicolumn{1}{l}{\multirow{2}{*}{BERT}} & $\mathcal{C}_{\rm B}$ & {75.6} & {74.4} & {59.4} & {59.9} & {\bf 53.9}  &{47.0} & \underline{\bf 77.5}  \\
        
        &  &  & $\mathcal{C}_{\rm B+Mix5}$ & {\bf 81.2} & {\bf 78.3} & {\bf 76.9} & \underline{\bf 72.4} & {53.0}  & \underline{\bf 56.8} & {75.8}  \\
        
        \midrule
        
        
         \multicolumn{1}{l}{\multirow{6}{*}{Few}} & \multicolumn{1}{l}{\multirow{4}{*}{PET}} & \multicolumn{1}{l}{\multirow{2}{*}{RoBERTa}} & $\mathcal{C}_{\rm B}$ & {88.6{\small ±1.5}} & {83.9{\small ±0.8}} & {87.8{\small ±0.7}} & {82.0{\small ±1.1}} & {82.8{\small ±5.6}} & {65.2{\small ±1.3}} & {86.0{\small ±0.4}}\\
        
        & & & $\mathcal{C}_{\rm R}$ & \underline{\bf 91.7{\small ±0.6}} & \underline{\bf 88.0{\small ±0.5}} & \underline{\bf 91.5{\small ±0.9}} & \underline{\bf 85.6{\small ±2.1}} & {\bf 87.8{\small ±2.2}} & \underline{\bf 68.9{\small ±1.0}} & \underline{\bf 87.8{\small ±0.9}} \\
        
        \cmidrule(lr){3-11}
        
        & & \multicolumn{1}{l}{\multirow{2}{*}{BERT}} & $\mathcal{C}_{\rm B}$ & {85.3{\small ±1.7}} & {80.3{\small ±2.1}} &	{89.2{\small ±0.3}} & {83.3{\small ±2.4}} & {85.4{\small ±1.9}} & {64.3{\small ±1.3}} & {84.0{\small ±1.0}} \\
        
        & & & $\mathcal{C}_{\rm B+Mix5}$ & {\bf 87.6{\small ±0.9}} & {\bf 85.0{\small ±0.8}} & {\bf 89.6{\small ±0.8}} & {\bf 85.0{\small ±1.7}} & {\bf 90.5{\small ±1.2}} & {\bf 68.4{\small ±0.7}} & \underline{\bf 87.8{\small ±0.6}} \\
        
         \cmidrule(lr){2-11}
        
        & \multicolumn{1}{l}{\multirow{2}{*}{NSP-BERT}} & \multicolumn{1}{l}{\multirow{2}{*}{BERT}} & $\mathcal{C}_{\rm B}$ & {86.7{\small ±2.1}} & {80.3{\small ±1.8}} & {86.7{\small ±1.7}} & {83.9{\small ±1.1}} & {86.6{\small ±0.9}}  & {64.5{\small ±0.5}} & {85.9{\small ±0.8}} \\
        
        &  &  & $\mathcal{C}_{\rm B+Mix5}$ & {\bf 89.4{\small ±0.7}} & {\bf 83.3{\small ±1.1}} & {\bf 88.7{\small ±1.0}} & {\bf 85.3{\small ±1.0}} &  \underline{\bf 92.1{\small ±1.1}} & {\bf 68.3{\small ±1.3}} & {\bf 87.6{\small ±0.5}}  \\
        \bottomrule
    \end{tabular}
    }
    \caption{Impact of pre-training corpus. $\mathcal{C}_{\rm B}$: pre-training from scratch with BERT's corpus; $\mathcal{C}_{\rm R}$: pre-training from scratch with RoBERTa's corpus; $\mathcal{C}_{\rm B+Mix5}$: continue pre-training with RoBERTa's corpus based on vanilla BERT.}
    \label{tb:corpus_matters}
    \vspace{-9pt}
\end{table*}

%% file: figures/fitting_process.tex
\begin{figure}[t]
\centering
\includegraphics[width=1.0\columnwidth]{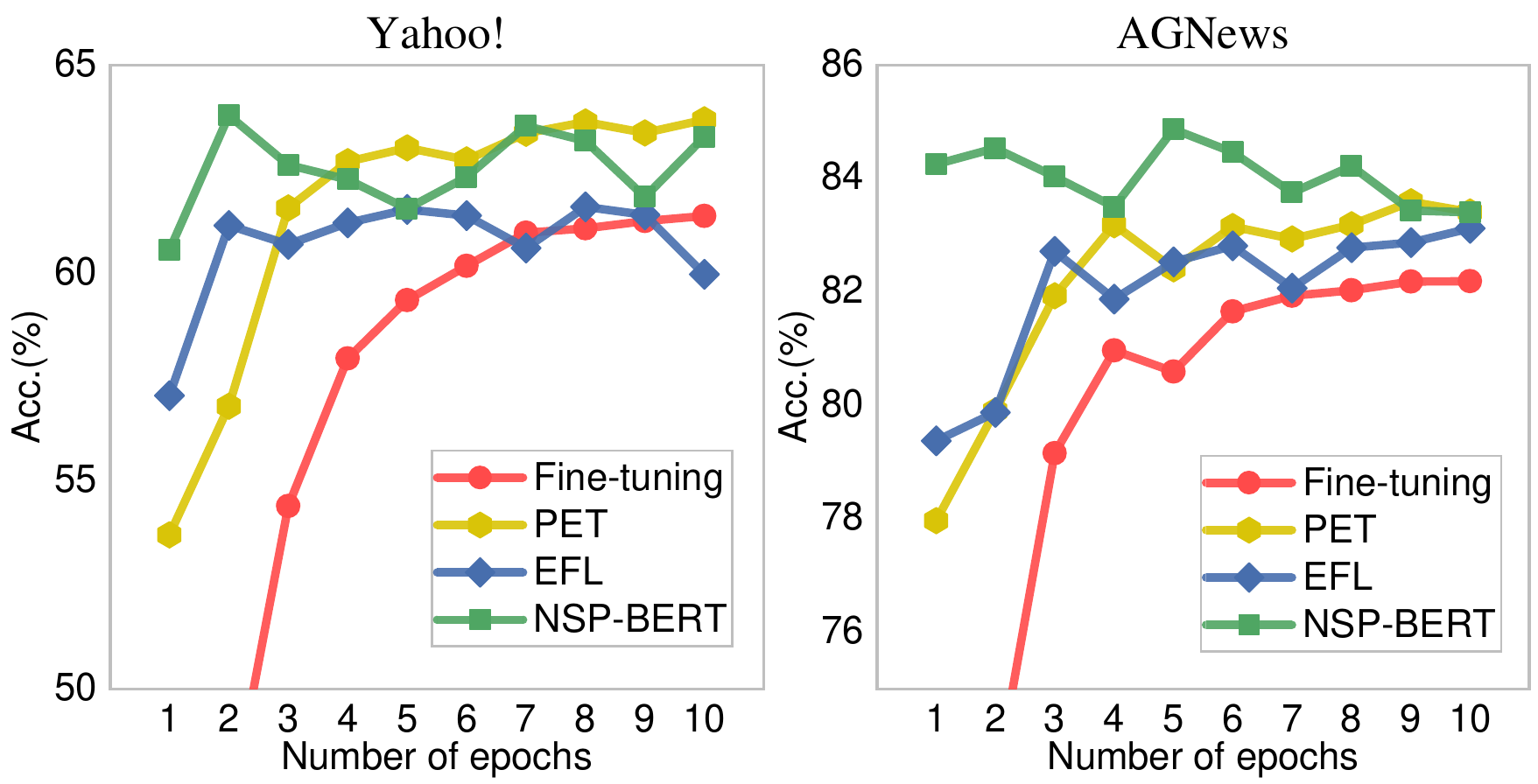} 
\caption{The accuracy of the 4 methods for each epoch during few-shot training on Yahoo! and AGNews.}
\label{fg:fitting_process}
\vspace{-6pt}
\end{figure}

%% file: tables/ablations.tex
\begin{table}[h]
    \begin{center}
    \centering
    \resizebox{1.0\columnwidth}{!}{%
    \begin{tabular}{lccccc}
    \toprule
    & {\bf SST-2} & {\bf MR}  &   {\bf CR}   & {\bf MPQA} \\
    \midrule
    NSP-BERT & {86.8{\small ±1.3}} & {\bf 80.5{\small ±1.5}} & {\bf 86.0{\small ±2.2}} & {\bf 83.9{\small ±1.1}}  \\
    \midrule
     \;coupled→decouple & {\bf 86.8{\small ±1.2}}  & {78.9{\small ±2.3}} & {85.8{\small ±9.5}} & {81.5{\small ±5.8}}  \\
    
    \;BCE→softmax & {83.8{\small ±5.0}} & {76.4{\small ±6.4}}  &  {80.5{\small ±10.0}} & {73.3{\small ±9.5}}  \\

    \;w/o NSP head & {83.8{\small ±6.5}} & {74.3{\small ±9.2}} & {79.0{\small ±8.1}} & {73.2{\small ±10.1}} \\
    
    \;linear head$+$softmax & {80.2{\small ±7.6}} & {71.9{\small ±12.3}} & {82.6{\small ±6.7}} & {73.8{\small ±11.1}} \\
    
    \bottomrule
    \end{tabular}
    }
    \end{center}
    \caption{Ablation studies of NSP-BERT on vanilla English BERT-Large. coupled→decouple: change coupled positive and negative samples to decoupled; BCE→softmax: change binary cross-entropy loss to softmax loss; w/o NSP head: use an initialized sigmoid head; linear head$+$softmax: use an initialized sigmoid head and softmax loss.}
    \vspace{-6pt}
    \label{tb:ablations}
\end{table}

%% file: figures/histogram.tex
\begin{figure*}[h]
\centering
\includegraphics[width=0.99\textwidth]{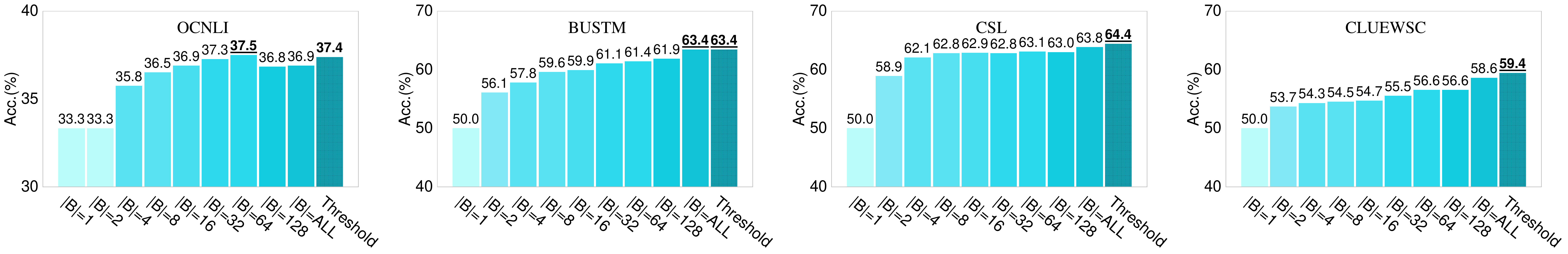} 
\caption{The performance of the samples-contrast answer mapping method under different preconditions on OCNLI, BUSTM, CSL and CLUEWSC. Batch size $|\mathcal{B}|\in \{1, 2,...,128, {\rm ALL}\}$, when the batch size is $1$ ($1$ and $2$ for OCNLI), the result is a random guess, when the batch size is ${\rm ALL}$, indicating that the entire test set is obtained at one time. {\tt Thresholds} means that the thresholds are obtained through the dev set, and then used for the prediction of the test set.}
\label{fg:histogram}
\vspace{-9pt}
\end{figure*}

%% file: figures/model_size.tex
\begin{figure}[t]
\centering
\includegraphics[width=1.0\columnwidth]{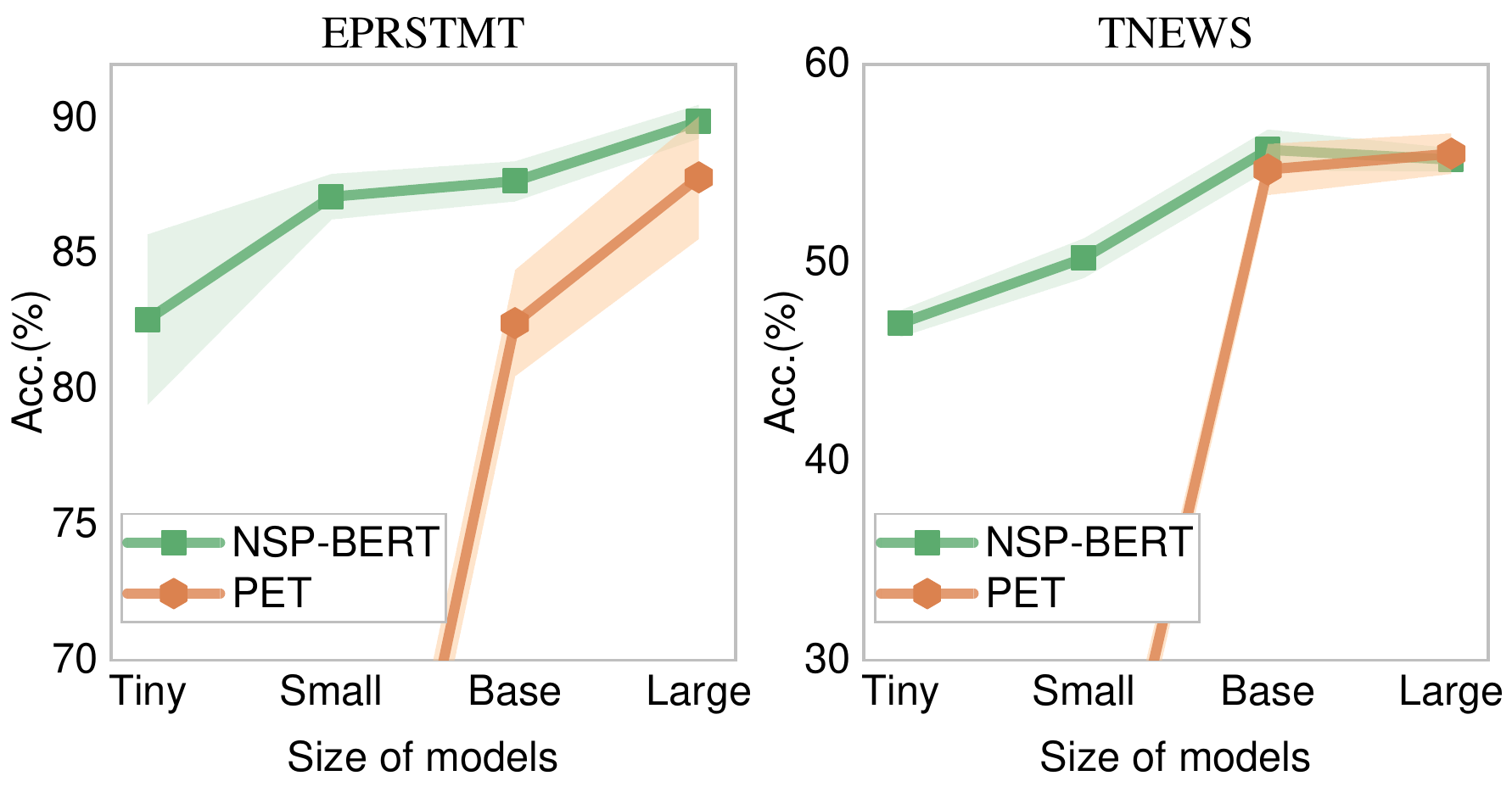} 
\caption{Accuracy of PET and NSP-BERT on EPRSTMT and TNEWS under 4 different model sizes.}
\label{fg:model_size}
\end{figure}

%% file: tables/applications.tex
\begin{table}[h]
    \begin{center}
    \centering
    \resizebox{1.0\columnwidth}{!}{%
    \begin{tabular}{lccccc}
    \toprule
    & \multicolumn{5}{c}{\multirow{1}{*}{\bf Chinese Tasks}}\\
    \cmidrule(lr){2-6}
    & {\bf OCNLI} & {\bf BUSTM} & {\bf CSL} & {\bf WSC} & {\bf ChID} \\
    \midrule
    Majority & {\it 38.1} & {\it 50.0} & {\it 50.0} & {\it 50.0} & {\it 14.3} \\
    PET & {\bf 40.3} & {50.6} & {52.2} & {54.7} & {\bf 57.6} \\
    NSP-BERT & {37.4} & {\bf 63.4} & {\bf 64.4} & {\bf 59.4} & {52.0} \\
    
    \midrule
    & \multicolumn{5}{c}{\multirow{1}{*}{\bf English Tasks}}\\
    \cmidrule(lr){2-6}
    & {\bf MNLI-m} & {\bf MNLI-mm} & {\bf SNLI} & {\bf QNLI} & {\bf RTE} \\
    \midrule
    Majority & {\it 32.7} & {\it 33.0} & {\it 33.8} & {\it 49.5} & {\it 52.7} \\
    PET & {\bf 47.1} & {\bf 46.0} & {36.0} & {49.0} & {51.6}  \\
    NSP-BERT & {39.4} & 39.2 & {\bf 43.4} & {\bf 67.6} & {\bf 55.6}  \\
  
    \bottomrule
    \end{tabular}
    }
    \end{center}
    \caption{Applications of NSP-BERT on FewCLUE tasks in zero-shot scenario. We report accuracy for all datasets. We only use the candidate-contrast method on ChID, and use the sample-contrast method on the rest of the datasets.}
    \vspace{-6pt}
    \label{tb:applications}
\end{table}

%% file: tables/two_stage_prompt.tex
\begin{table}[h]
    \begin{center}
    \centering
    \resizebox{0.75\columnwidth}{!}{%
    \begin{tabular}{lcc}
    \toprule
    & {\bf DuEL2.0-L} & {\bf DuEL2.0-T}  \\
    \midrule
    
    PET & {-} & 40.0 \\
    NSP-BERT & {61.2} / {69.7↑}  & {31.4} / {40.0↑} \\
  
    \bottomrule
    \end{tabular}
    }
    \end{center}
    \caption{Word sense disambiguation task. DuEL2.0-L: DuEL2.0 entity linking part; DuEL2.0-T: DuEL2.0 entity typing part. The left side of the slash is the one-stage prompt, and the right side is the two-stage prompt.}
    \vspace{-6pt}
    \label{tb:two_stage_prompt}
\end{table}

%% file: sections/05_conclusion.tex
\section{Conclusion}

In this paper, we show that NSP can also be an apposite zero-shot or few-shot learner same as MLM. This not only provides a new route for prompt-learning, but also makes us rethink the role of sentence-level pre-training tasks. At the same time, we continue to pre-train the BERT model with a small amount of computing power, and its performance improves significantly on both zero-shot and few-shot learning, whether to use PET or NSP-BERT. We believe that not only the size of the model, but also the pre-training corpus, both determine the upper limit of the model's ability on few-shot learning.

%% file: Appendix.tex
\section{Models}

\subsection{Probability Formula}
We compared the output probability formulas of different zero-shot prompt-learning models include our NSP-BERT. The following description is a general situation, assuming that each label it mapped to a span with a length is greater than or equal to $1$. When the length of the label word is equal to 1, the form of the pre-training and downstream tasks tend to be unified. When the length is greater than 1, there is a gap between them, even we use the model pre-trained by whole word masking \cite{cui2019pre} or span masking \cite{joshi2020spanbert}.
\label{sec:probs}
\paragraph{PET-{\sc Zero}} Denote the token in position $i$ as $t_i$, the label span will be replaced by $\mask_{l:r}$. When ignoring special tokens such as \cls and \pad, the input of PET-{\sc Zero} is:
\begin{equation}
\resizebox{.92\hsize}{!}{$\displaystyle
    \mathbf{x}_{input}=t_1,...,\mask _l,...,\mask _r,...
$}
\end{equation}
The output probability for label $y_{i}^{( j )}$ is:
\begin{equation}
\resizebox{.89\hsize}{!}{$\displaystyle
    q( y_{i}^{( j )}|\mathbf{x}_{i} ) =\underset{1\leqslant j\leqslant M}{\rm softmax}(\prod_{l\leqslant v\leqslant r}{q_{\mathcal{M} _{\mathrm{MLM}}}(t_{v}^{_{\left( j \right)}}|\mathbf{x}_{input}})).
$}   
\end{equation}

\paragraph{NSP-BERT} For our NSP-BERT, the label span $t^{(j)}_{l:r}$ will be replaced in turn:
\begin{equation}
    \mathbf{x}_{input}^{(j)}=t_1,...,\sep ,...,t_{l}^{( j )},...,t_{r}^{( j )},...
\end{equation}
The output probability for label $y_{i}^{( j )}$ is:
\begin{equation}
    q( y_{i}^{( j )}|\mathbf{x}_{i} ) =\underset{1\leqslant j\leqslant M}{\rm softmax}(q_{\mathcal{M} _{\mathrm{NSP}}}(\mathbf{x}_{input}^{(j)})).
\end{equation}

\subsection{Parameters of Models}
For FewCLUE, we use the Chinese vanilla-BERT-{\sc Base} pre-trained by UER \cite{zhao2019uer} for the main results of our NSP-BERT. We also report the results of the other scales (tiny, small and large) model. Following the implementation of \cite{xu2021fewclue}, we use Chinese RoBERTa-wwm-ext-{\sc Base} pre-trained by HFL \cite{cui2019pre} and NEZHA-Gen \cite{wei2019nezha} for the baselines.

For English datasets, following the implementation \footnote{https://github.com/princeton-nlp/LM-BFF} of \cite{gao2021making}. We use vanilla-BERT-{\sc Large} pre-trained by Google \cite{devlin2018bert} for our NSP-BERT, and RoBERTa-{\sc Large}\footnote{https://github.com/pytorch/fairseq/tree/main/examples/roberta} for the baselines.

Table \ref{tb:model_paramaters} shows the hyperparameters of the models used in our experiment. The English and Chinese models are a little different in total parameters, mainly due to the different vocabulary size.
It should be noted that not all pre-trained models fully stored NSP head and MLM head, so we need to select deliberately.

\input{tables/models}

\subsection{Others}
\begin{figure*}[h]
\centering
\includegraphics[width=1.0\textwidth]{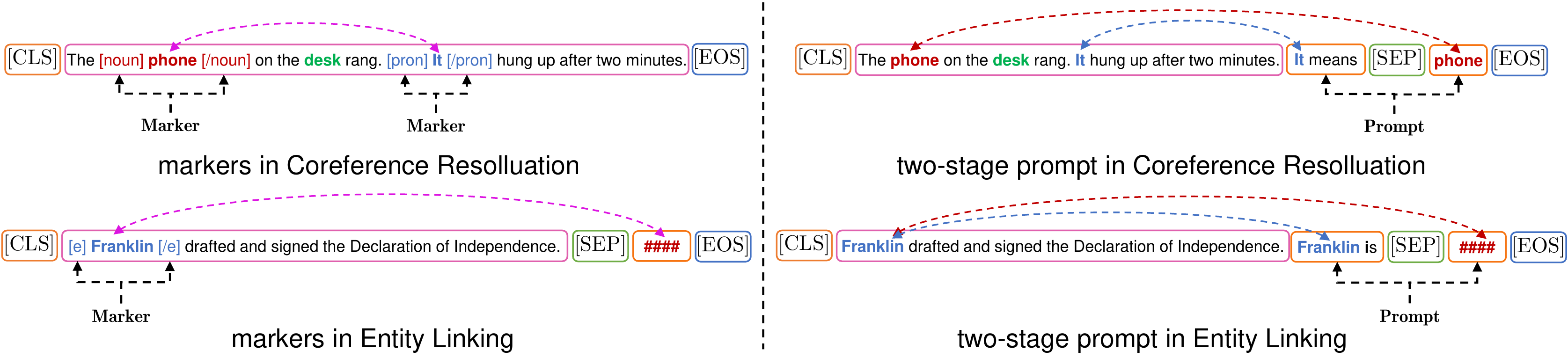} 
\caption{The comparison of markers (Left) and two-stage prompt (Right), examples in coreference resolution and entity linking/typing tasks.}
\label{fg:two_stage_compare}
\end{figure*}

\paragraph{Marks and Two-stage prompt}
In the Figure \ref{fg:two_stage_compare}, we compare the markers that usually appear in supervised training \cite{huang2019glossbert, soares2019matching, wu2019enriching, zhong2021a}. The marker are special tokens such as \noun, \pron and \e. They are usually added before and after the target words. The two-stage prompt plays the same role as the markers, but it uses a natural language description method.
\section{More Details}

\subsection{Datasets}

\label{sec:appendix_datasets}

\paragraph{FewCLUE} FewCLUE \cite{xu2021fewclue} is a Chinese few-shot learning evaluation benchmark with 9 Chinese NLU tasks in total. There are 4 single-sentence tasks which are EPRSTMT, TNEWS, CLSDCP and IFLYTEK.  EPRSTMT is a binary sentiment analysis dataset for E-commerce reviews. TNEWS \cite{xu2020clue} is a short text classification for news title with 15 topics. CSLDCP is a text classification dataset including abstracts from a variety of Chinese scientific papers and with 67 categories in total. IFLYTEK \cite{2019iflytek} is a long text classification dataset for App descriptions. There are 3 sentence-pair tasks which are OCNLI, BUSTM and CSL. OCNLI \cite{ocnli} is an original Chinese NLI tasks. BUSTM \cite{BUSTM} is a dialogue short text matching task. CSL is a abstract-keywords matching task. There are other two tasks ChID and CLUEWSC. ChID \cite{zheng2019chid} is a Chinese idiom cloze test dataset. CLUEWSC is a coreference resolution task.

For all the datasets in FewCLUE, we evaluate our model on the public test set. Although FewCLUE provides a large number of unlabeled samples, we did not use them in the our experiment, so the results are unable to be compared with the results on the leaderboard\footnote{https://www.cluebenchmarks.com/fewclue.html}. For dataset TNEWS, we did not use the information of keywords following \cite{xu2021fewclue}. We treat CLUEWSC as a sentence-pair task due to its data characteristics.

\paragraph{DuEL2.0}
We divide DuEL2.0 into two parts. In the first part, the entity linking part, there are 26586 samples. All the samples' mention can be mapped to single or multiple entities in the knowledge base, and each mention can be linked to 5.37 entities on average. In the second part, the entity typing part, there are 6465 samples. Those samples' mention cannot be found in the knowledge base, but they will be divided into their corresponding upper entity types. There are a total of 24 upper entity types, and we do not remove the {\tt Other} type. When performing the entity linking part, we only use the entity's summary information, without using more entity triples.
\input{tables/duel}
\input{tables/task_details}
\paragraph{English Datasets}
Following \cite{gao2021making,hu2021knowledgeable,liang2022contrastive}, we evaluate our model on 7 single-sentence and 5 sentence-pair English tasks. For the datasets SST-2 \cite{socher2013recursive_sst-2}, MNLI \cite{williams2018broad_mnli}, QNLI \cite{rajpurkar2016squad}, RTE \cite{dagan2005pascal_rte1, bar2006second, giampiccolo2007third_rte3, bentivogli2009fifth_rte4}, we follow \cite{gao2021making} and \cite{zhang2021revisiting} and use their original development sets for testing. For datasets MR \cite{pang2005seeing_mr}, CR \cite{hu2004mining_cr}, MPQA \cite{wiebe2005annotating_mpqa}, Subj \cite{pang2004sentimental_subj}, Yahoo! and AGNews\cite{zhang2015character}, we use the testing set randomly sampled from training set and leaved from training by \cite{gao2021making}\footnote{https://nlp.cs.princeton.edu/projects/lm-bff/datasets.tar}. For SNLI \cite{bowman2015large_snli}, we use their official test sets. 

\subsection{Results}


\paragraph{Different Templates} 
We compared in detail the performance of NSP-BERT under different prompt templates. This experiment wad conducted on 4 Chinese single-sentence classification datasets. 
\begin{compactitem}
\item {\bf Template 1} uses just the original label words.
\item {\bf Template 2} adds pronouns and copulas such as ``I am'', ``it is'' or ``this is'', to make the template become a complete sentence.
\item {\bf Template 3} incorporates more domain information into the prompts, such as ``shopping'', ``news'', ``paper'' and ``app''. This makes the original input sentence and prompt have better connectivity.
\end{compactitem}
For zero-shot learning, the prompt templates have a strong impact on the performance, and for different models, there is a big difference. Therefore, we verified the influence of templates for different models versions and scales. The results are shown in Table \ref{tb:eprstmt}, Table \ref{tb:tnews}, Table \ref{tb:csldcp} and Table \ref{tb:iflytek}.

\input{tables/eprstmt}
\input{tables/tnews}
\input{tables/csldcp}
\input{tables/iflytek}

\begin{figure*}[h]
\centering
\includegraphics[width=0.85\textwidth]{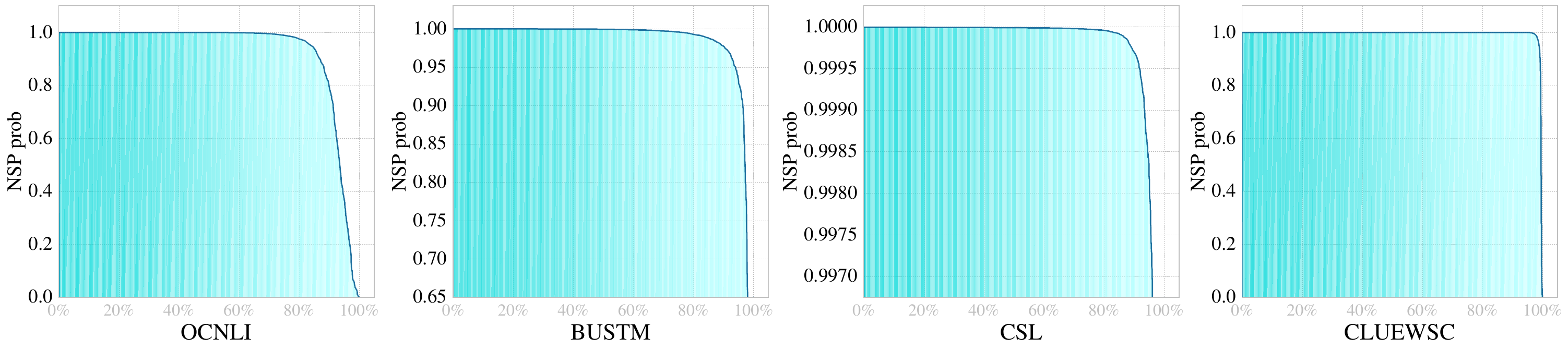} 
\caption{The NSP output probability of the 4 sentence-pair tasks OCNLI, BUSTM, CSL and CLUEWSC in Chinese benchmark FewCLUE. The x-axis represents the proportion of the samples. And the y-axis represents the NSP probability of the samples.}
\label{fg:probas_zh}
\end{figure*}

\begin{figure*}[h]
\centering
\includegraphics[width=0.85\textwidth]{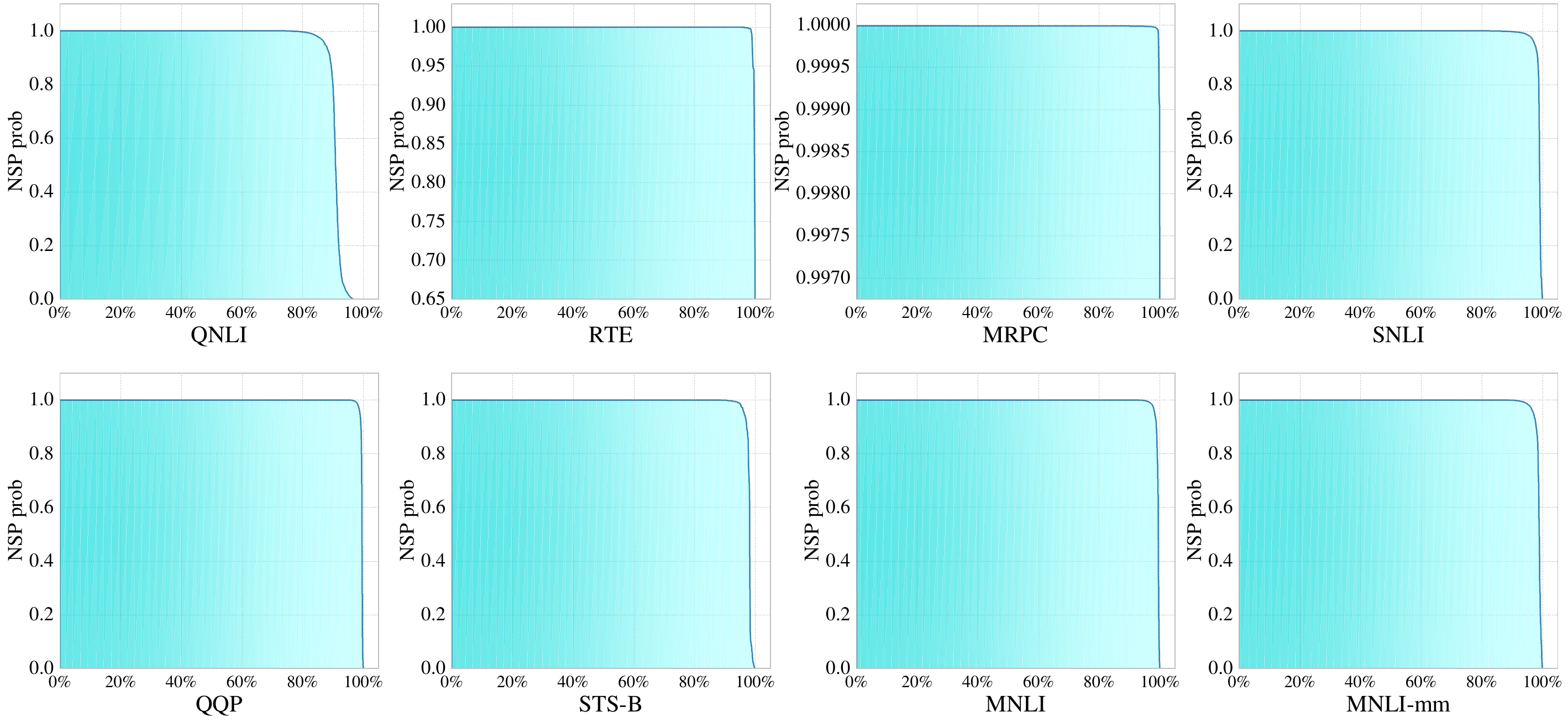} 
\caption{The NSP output probability of the 8 English  sentence-pair tasks QNLI, RTE, MRPC, SNLI, QQP, STS-B, MNLI and MNLI-mm. The x-axis represents the proportion of the samples. And the y-axis represents the NSP probability of the samples.}
\label{fg:probas_en}
\end{figure*}

\paragraph{Probability of NSP in sentence-pair tasks}
\label{sec:prob_sentence_pair}
To further explain the necessity for us to propose sample-contrast mapping method, we show the NSP output probability of the sentence-pair tasks in Figure \ref{fg:probas_zh} and Figure \ref{fg:probas_en}. It's not difficult to see that the NSP probability of most samples is close to $1$. So we can not judge its label for a individual sample. We need to contrast different samples, and predict the label by obtaining the distribution of the dataset.

\paragraph{Impact of batch size for samples-contrast}
In one case, we cannot get the entire test set at once, then we need to predict the samples of the test set batch by batch. We set the batch size $|B|\in \{1, 2,...,128, {\rm ALL}\}$, to observe the results predicted by samples-contrast method (see Table \ref{tb:samples_contrast}). As the batch size increases, the performance improves and stabilizes. Of course, when the batch size is less than the number of labels, the result is equivalent to random guessing. In another case, we cannot get the distribution of the test set, that is, we don't know the proportion of each label. Then we can use the development to calculate the NSP probability threshold of each label to predict the test set. The model can also get the desired performance.

\paragraph{Strategies for datasets}
For different datasets, according to their characteristics, the position of the prompt (prefix or suffix), and the mapping method (candidates-contrast or samples-contrast) are different. We take Chinese tasks as examples, all the strategies are shown in Table \ref{tb:strategies}. In the single-sentence classification tasks (EPRSTMT, TNEWS, CSLDCP, IFLYTEK), the prompts are all prefixed, and we adopt candidates-contrast. For the word sense disambiguation tasks (CLUEWSC and DuEL2.0), since we need to utilize two-stage prompt method, we all use the suffix. In sentence-pair tasks (OCNLI, BUSTM and CSL), we choose the appropriate order through the development set to arrange the two sentences, where suffix means using the original order and prefix means using the reverse order.

\paragraph{Prompts for datasets}
Due to the number of data sets in our paper, we report in detail the prompt templates of the more important Chinese datasets in Table \ref{tb:prompts}, and briefly report the prompts of English datasets in Table \ref{tb:prompts_en}.





\input{tables/samples_contrast}

\input{tables/strategies}

\input{tables/prompts_en}

\input{tables/prompts}

%% file: tables/models.tex
\begin{table}[h]
    \centering
    \renewcommand\arraystretch{1.0}
    \resizebox{1\columnwidth}{!}{
    \begin{tabular}{lccccc}
    \toprule
        {\bf Model} & $\boldsymbol{L}$ & $\boldsymbol{H}$ & $\boldsymbol{A}$  & \multicolumn{2}{c}{\makecell[c]{\bf Total Parameters \\ ZH \quad/\quad EN}} \\ \midrule
        {\bf RoBERTa} & 12 & 768 & 12 & 102M& - \\
        {\bf RoBERTa-{\sc Large}} & 12 & 768 & 12 & - & 355M \\ \midrule
        {\bf BERT-{\sc Tiny}} & 3 & 384 & 6 & 14M & -\\ 
        {\bf BERT-{\sc Small}} & 6 & 512 & 8 & 31M &  -\\ 
        {\bf BERT-{\sc Base}} & 12 & 768 & 12 & 102M & -\\ 
        {\bf BERT-{\sc Large}} & 24 & 1024 & 16 & 327M & 355M \\ 
    \bottomrule
    \end{tabular}
    }
    \caption{The parameters of different models used in our experiment. $L$: number of layers; $H$: hidden size; $A$: number of self-attention heads; ``-'': not used in our paper; ZH: Chinese model; EN: English model.}
    \label{tb:model_paramaters}
\end{table}

%% file: tables/duel.tex
\begin{table}[h]
    \centering
    \renewcommand\arraystretch{1.1}
    \resizebox{0.99\columnwidth}{!}{
    \begin{tabular}{cc|cc}
    \toprule
        Entity Linking & Ave. Entities & Entity Tpying & Types \\ \midrule
        26586 & 5.37 & 6465 & 24 \\ 
    \bottomrule
    \end{tabular}
    }
    \caption{Since the DuEL2.0's test set is not public, we use the dev set to test our model. The the number of the original text lines is 10000. According to the predicted target (entities in knowledge base or upper types), we manually divide it into two parts, entity linking and entity typing.}
    \label{tb:duel}
\end{table}

%% file: tables/task_details.tex
\begin{table*}[h]
    \centering
    \resizebox{1\textwidth}{!}{
    \begin{tabular}{llrrcccl}
    \toprule
        \textbf{Category} & \textbf{Corpus} & \textbf{\#Train}& \textbf{\#Test} & \textbf{$|\mathcal{Y}|$} & \textbf{Task Type} & \textbf{Metrics} & \textbf{Source} \\ \midrule
       
        \multicolumn{8}{c}{\textbf{English Tasks}} \\  
        \midrule
        & SST-2 & 6,920 & 872 & 2 & Sentiment Analysis & Acc. & Movie Reviews \\
        & MR & 8,662 & 2,000 & 2 & Sentiment Analysis & Acc. & Movie Reviews \\
        Single-   & CR & 1,775 & 2,000 & 2 & Sentiment Analysis & Acc. & E-commerce Reviews \\
        Sentence  & MPQA & 8,606 & 2,000 & 2 & Opinion Polarity & Acc. & World Press \\
        & Subj & 8,000 & 2,000 & 2 & Subjectivity & Acc. & Movie Reviews \\
        & Yahoo! & 1,400,000  & 6,000 & 10 & Question Classification & Acc. & Yahoo \\
        & AGNews & 8,551 & 7,600 & 4 & News Topic Classification & Acc. & Web \\
        
        \midrule
        & MNLI    & 392,702  & 9,815 & 3 & Natural Language Inference & Acc. &  Speech, Fiction and Reports \\
        & MNLI-mm & 392,702 & 9,832 & 3 & Natural Language Inference & Acc. & Speech, Fiction and Reports \\
        Sentence- & SNLI    & 549,367 & 9,842 & 3 & Natural Language Inference & Acc. & Image Captions \\
        Pair & QNLI    & 104,743 & 5,463 & 2 & Natural Language Inference & Acc. & Wikipedia \\
        & RTE  & 2,490 & 277 & 2 & Natural Language Inference & Acc. & News and Wikipedia \\
        
        \midrule
         \multicolumn{8}{c}{\textbf{Chinese Tasks (FewCLUE)}} \\  
        \midrule
        &{EPRSTMT} & 32 & 610 & 2 & Sentiment Analysis & Acc. & E-commerce Reviews \\ 
        Single- & {TNEWS} & 240 & 2,010 & 15 & Short Text Classification & Acc. & News Title \\ 
        Sentence & {CSLDCP} & 536 & 1,784 & 67 & Long Text Classification & Acc. & Academic CNKI \\ 
        & {IFLYTEK} & 928 & 1,749 & 119 & Long Text Classification & Acc. & App Description \\   \midrule
        Sentence- & {OCNLI} & 32 & 2,520 & 3 & Natural Language Inference & Acc. & 5 genres  \\ 
        Pair & {BUSTM} & 32 & 1,772 & 2 & Short Text Matching & Acc. & AI Virtual Assistant \\  
        & {CSL} & 32 & 2,828 & 2 & Keyword Recognition & Acc. & Academic CNKI \\ \midrule
        Others & {ChID} & 42 & 2,002 & 7 & Chinese Idiom Cloze Test & Acc. & Novel, Essay News \\ 
        & {CLUEWSC} & 32 & 976 & 2 & Coreference Resolution & Acc. & Chinese Fiction Books \\

    \bottomrule
    \end{tabular}
    }
    \caption{
    Task descriptions and statistics. In FewCLUE we omit the unlabeled dataset because it is not used. Test of FewCLUE indicates the number of samples in the public test set. The 5 text genres of OCNLI are government documents, news, literature, TV talk shows and telephone conversations. 
    }
    \label{tb:datasets}
\end{table*}

%% file: tables/eprstmt.tex
\begin{table}[h]
    \center
    \renewcommand\arraystretch{1.0}
    \resizebox{0.99\columnwidth}{!}{
    \begin{tabular}{llccc}
    \toprule
        \multirow{2}{*}{\bf ORG} & \multicolumn{1}{c}{\multirow{2}{*}{\bf Models}} & {\bf Template 1} & {\bf Template 2} & {\bf Template 3} \\
         &  & (Dev/Test) & (Dev/Test) & (Dev/Test) \\ 
         \midrule
        \multicolumn{1}{l}{\multirow{4}{*}{\bf UER}} & \multicolumn{1}{l}{\small BERT-{\sc Tiny}} & 68.13/76.56 & 75.00/80.82 & {\bf 81.88/80.33} \\ 
        \multicolumn{1}{l}{} & \multicolumn{1}{l}{\small BERT-{\sc Small}} & 85.00/87.70 & 82.50/87.70 & {\bf 87.50/86.72} \\ 
        \multicolumn{1}{l}{} & \multicolumn{1}{l}{\small BERT-{\sc Base}} & 60.00/54.59 & 78.75/80.98 & {\bf \underline{88.13/86.89}} \\ 
        \multicolumn{1}{l}{} & \multicolumn{1}{l}{\small BERT-{\sc Large}} & 78.13/82.79 & 83.75/82.62 & {\bf 84.38/84.43} \\ 
        
        \bottomrule
    \end{tabular}
    }
    \caption{Zero-shot acc. of NSP-BERT on EPRSTMT. }
    \label{tb:eprstmt}
\end{table}

%% file: tables/tnews.tex
\begin{table}[h]
    \center
    \renewcommand\arraystretch{1.0}
    \resizebox{0.99\columnwidth}{!}{
    \begin{tabular}{llccc}
    \toprule
        \multirow{2}{*}{\bf ORG} & \multicolumn{1}{c}{\multirow{2}{*}{\bf Models}} & {\bf Template 1} & {\bf Template 2} & {\bf Template 3} \\
         &  & (Dev/Test) & (Dev/Test) & (Dev/Test) \\ 
         \midrule
        \multicolumn{1}{l}{\multirow{4}{*}{\bf UER}} & \multicolumn{1}{l}{\small BERT-{\sc Tiny}} & 38.80/36.62 & 39.25/36.37 & {\bf 41.07/38.56} \\ 
        \multicolumn{1}{l}{} & \multicolumn{1}{l}{\small BERT-{\sc Small}} & 38.98/38.81 & 39.80/40.35 & {\bf 41.80/42.19} \\
        \multicolumn{1}{l}{} & \multicolumn{1}{l}{\small BERT-{\sc Base}} & 41.26/41.84 & 46.99/48.66 & {\bf 50.64/51.00}\\ 
        \multicolumn{1}{l}{} & \multicolumn{1}{l}{\small BERT-{\sc Large}} & 45.17/42.79 & 48.72/48.31 & {\bf \underline{54.28/53.83}} \\ 
        
        \bottomrule
    \end{tabular}
    }
    \caption{Zero-shot acc. of NSP-BERT on TNEWS. }
    \label{tb:tnews}
\end{table}

%% file: tables/csldcp.tex
\begin{table}[h]
    \center
    \renewcommand\arraystretch{1.0}
    \resizebox{0.99\columnwidth}{!}{
    \begin{tabular}{llccc}
    \toprule
        \multirow{2}{*}{\bf ORG} & \multicolumn{1}{c}{\multirow{2}{*}{\bf Models}} & {\bf Template 1} & {\bf Template 2} & {\bf Template 3} \\
         &  & (Dev/Test) & (Dev/Test) & (Dev/Test) \\ 
         \midrule
        \multicolumn{1}{l}{\multirow{4}{*}{\bf UER}} & \multicolumn{1}{l}{\small BERT-{\sc Tiny}} & 24.03/25.73 & {\bf 27.37/29.60} & 25.68/28.81 \\ 
        \multicolumn{1}{l}{} & \multicolumn{1}{l}{\small BERT-{\sc Small}} & 28.48/30.72 & 29.35/31.45 & {\bf 29.78/31.78} \\
        \multicolumn{1}{l}{} & \multicolumn{1}{l}{\small BERT-{\sc Base}} & 39.80/40.53 & 44.87/45.80 & 45.26/{\bf \underline{47.59}} \\ 
        \multicolumn{1}{l}{} & \multicolumn{1}{l}{\small BERT-{\sc Large}} & 44.73/42.83 & 44.00/44.34 & {\bf \underline{45.89}}/46.92 \\ 
        
        \bottomrule
    \end{tabular}
    }
    \caption{Zero-shot acc. of NSP-BERT on CSLDCP. }
    \label{tb:csldcp}
\end{table}

%% file: tables/iflytek.tex
\begin{table}[h]
    \center
    \renewcommand\arraystretch{1.0}
    \resizebox{0.99\columnwidth}{!}{
    \begin{tabular}{llccc}
    \toprule
        \multirow{2}{*}{\bf ORG} & \multicolumn{1}{c}{\multirow{2}{*}{\bf Models}} & {\bf Template 1} & {\bf Template 2} & {\bf Template 3} \\
         &  & (Dev/Test) & (Dev/Test) & (Dev/Test) \\ 
         \midrule
        \multicolumn{1}{l}{\multirow{4}{*}{\bf UER}} & \multicolumn{1}{l}{\small BERT-{\sc Tiny}} & 32.70/32.65 & 31.97/34.13 & {\bf 33.65/34.59} \\ 
        \multicolumn{1}{l}{} & \multicolumn{1}{l}{\small BERT-{\sc Small}} & 32.27/32.42 & {\bf 35.54}/34.65 & 35.25/{\bf 34.76} \\
        \multicolumn{1}{l}{} & \multicolumn{1}{l}{\small BERT-{\sc Base}} & 36.41/36.59 & 42.39/40.19 & {\bf 43.12/41.62} \\ 
        \multicolumn{1}{l}{} & \multicolumn{1}{l}{\small BERT-{\sc Large}} & 37.73/36.94 & 44.28/{\bf \underline{42.60}} & {\bf \underline{44.87}}/42.42 \\ 
        
        \bottomrule
    \end{tabular}
    }
    \caption{Zero-shot acc. of NSP-BERT on IFLYTEK. }
    \label{tb:iflytek}
\end{table}

%% file: tables/samples_contrast.tex
\begin{table*}[h]
    \centering
    \resizebox{0.99\textwidth}{!}{
    \begin{tabular}{lccccccccccc}
    \toprule
        \multirow{2}{*}{\bf Dataset} & \multirow{2}{*}{\bf Dev} &  &  &  &  &  & {\bf Test} &  &  &  &  \\ \cmidrule(r){3-12}
         &  & $|\mathcal{B}|=$1 & $|\mathcal{B}|=$2 & $|\mathcal{B}|=$4 & $|\mathcal{B}|=$8 & $|\mathcal{B}|=$16 & $|\mathcal{B}|=$32 & $|\mathcal{B}|=$64 & $|\mathcal{B}|=$128 & $|\mathcal{B}|=$All & Threshold \\ \midrule
        \multicolumn{1}{l|}{OCNLI}  & \multicolumn{1}{l|}{37.50} & 33.33 & 33.33 & 35.75 & 36.51 & 36.90 & 37.26 & \underline{\bf 37.50} & 36.83 & 36.90 & 37.38 \\ 
        \multicolumn{1}{l|}{BUSTM}  & \multicolumn{1}{l|}{62.50} & 50.00 & 56.09 & 67.79 & 59.59 & 59.93 & 61.06 & 61.40 & 61.85 & \underline{\bf 63.43} & \underline{\bf 63.43} \\ 
        \multicolumn{1}{l|}{CSL}  & \multicolumn{1}{l|}{64.38} & 50.00 & 58.91 & 62.09 & 62.79 & 62.86 & 62.79 & 63.07 & 63.00 & 63.85 & \underline{\bf 64.41} \\ 
        \multicolumn{1}{l|}{CLUEWSC}  & \multicolumn{1}{l|}{57.23} & 50.00 & 53.69 & 54.30 & 54.51 & 54.71 & 55.53 & 56.56 & 56.56 & 58.61 & \underline{\bf 59.43} \\ 
        \midrule
        \multicolumn{1}{l|}{MNLI-m}  & \multicolumn{1}{l|}{41.67} & 35.22 & 35.22 & 39.08 & \underline{\bf 40.04} & 39.08 & 39.63 & 39.33 & 39.48 & 39.33  &  39.41\\ 
        \multicolumn{1}{l|}{MNLI-mm}  & \multicolumn{1}{l|}{39.58} & 35.45 & 35.45 & 38.41 & 38.59 & 38.62 & 38.19 & 37.69 & 38.24 & 38.17 & \underline{\bf 39.17}\\ 
        \multicolumn{1}{l|}{SNLI}  & \multicolumn{1}{l|}{43.75} & 34.28 & 34.28 & \underline{\bf 44.14} & 44.21 & 43.54 & 43.20 & 43.17 & 43.13 & 43.35 & 43.42 \\ 
        \multicolumn{1}{l|}{QNLI}  & \multicolumn{1}{l|}{87.50} & 49.46 & 62.37 & 64.63 & 65.37 & 66.58 & 66.87 & 67.23 & 67.34 & 67.56 & \underline{\bf 67.56} \\ 
        \multicolumn{1}{l|}{RTE}  & \multicolumn{1}{l|}{62.50} & 52.71 & 52.71 & 54.87 & 53.43 & 55.60 & 54.15 & \underline{\bf 54.15} & 54.87 & 51.99 & 55.60 \\ 
    \bottomrule
    \end{tabular}
    }
    \caption{The performance of the samples-contrast answer mapping method under different preconditions on sentence-pair tasks. Batch size $|\mathcal{B}|\in \{1, 2,...,128, {\rm ALL}\}$, when the batch size is less than the number of labels, the result is a random guess, when the batch size is ${\rm ALL}$, indicating that the entire test set is obtained at one time. {\tt Thresholds} means that the thresholds are obtained through the development set, and then used for the prediction of the test set.}
    \label{tb:samples_contrast}
\end{table*}

%% file: tables/strategies.tex
\begin{table*}[h]
    \centering
    \renewcommand\arraystretch{1.0}
    \resizebox{1.0\textwidth}{!}{
    \begin{tabular}{lccccc|ccc|cc|cc}
    \toprule
        \multicolumn{2}{c|}{\multirow{2}{*}{\bf Strategies}} &\multicolumn{4}{c|}{\bf Single-Sentence Task} &\multicolumn{3}{c|}{\bf  Sentence-Pair Task} & \multicolumn{2}{c}{\bf Others} & \multicolumn{2}{|c}{\bf DuEL2.0} \\ \cmidrule(r){3-6} \cmidrule(r){7-9}  \cmidrule(r){10-13}
        & \multicolumn{1}{c|}{} & {\small EPRSTMT} & {\small TNEWS} & {\small CSLDCP} & {\small IFLYTEK} & {\small OCNLI} & {\small BUSTM} & {\small CSL} & {\small ChID} & {\small CLUEWSC} & {\small Entity Linking} & {\small Entity Typing}\\ \midrule
        \multicolumn{1}{c|}{\multirow{2}{*}{\bf Prompt}} & \multicolumn{1}{l|}{Prefix} & \checkmark & \checkmark & \checkmark & \checkmark & \checkmark & \checkmark &  &  &  & & \\ \cmidrule(r){2-13}
        \multicolumn{1}{c|}{} & \multicolumn{1}{l|}{Suffix} &  &  &  &  &  &  & \checkmark & \checkmark & \checkmark & \checkmark & \checkmark\\ \midrule
        \multicolumn{1}{c|}{\bf Answer} & \multicolumn{1}{c|}{C-C} & \checkmark & \checkmark & \checkmark & \checkmark &  &  &  & \checkmark &  & \checkmark & \checkmark \\ \cmidrule(r){2-13}
        \multicolumn{1}{c|}{\bf Mapping} & \multicolumn{1}{c|}{S-C} &  &  &  &  & \checkmark & \checkmark & \checkmark &  & \checkmark & & \\ 
    \bottomrule
    \end{tabular}
    }
    \caption{Strategies adopted on the 10 datasets in FewCLUE and DuEL2.0. The {\bf prefix} means to put the prompt in front of the original text, and the {\bf suffix} is the opposite. {\bf C-C} means candidates-contrast answer mapping method, and {\bf S-C} means samples-contrast answer mapping method.}
    \label{tb:strategies}
\end{table*}

%% file: tables/prompts_en.tex
\begin{table*}[h]
    \begin{center}
    \centering
    \resizebox{0.7\textwidth}{!}{%
    \begin{tabular}{llll}
    \toprule
    {\bf Task} & {\bf Method} & {\bf Prompt Templates}\\
    \midrule
    
    \multicolumn{1}{l}{\multirow{5}{*}{{\bf SST-2}}} & & {Original Labels}:~~~~negative; positive \\
    \cmidrule(lr){2-3}
    &  \multicolumn{1}{l}{\multirow{2}{*}{PET}} & {Mapping Words}:~~~terrible; great \\
    & & {Prompt Template}:~{\sent} It was {\lab}. \\
    \cmidrule(lr){2-3}
    &  \multicolumn{1}{l}{\multirow{2}{*}{NSP-BERT}} & {Mapping Words}:~~~terrible; great \\
    & & {Prompt Template}:~A {\lab} piece of work {\sep} {\sent}  \\
    \midrule
    
    \multicolumn{1}{l}{\multirow{5}{*}{{\bf MR}}} & & {Original Labels}:~~~~negative; positive \\
    \cmidrule(lr){2-3}
    &  \multicolumn{1}{l}{\multirow{2}{*}{PET}} & {Mapping Words}:~~~terrible; great \\
    & & {Prompt Template}:~{\sent} It was {\lab}. \\
    \cmidrule(lr){2-3}
    &  \multicolumn{1}{l}{\multirow{2}{*}{NSP-BERT}} & {Mapping Words}:~~~terrible; great \\
    & & {Prompt Template}:~A {\lab} piece of work {\sep} {\sent}  \\
    \midrule
    
    \multicolumn{1}{l}{\multirow{5}{*}{{\bf CR}}} & & {Original Labels}:~~~~negative; positive \\
    \cmidrule(lr){2-3}
    &  \multicolumn{1}{l}{\multirow{2}{*}{PET}} & {Mapping Words}:~~~terrible; great \\
    & & {Prompt Template}:~{\sent} It was {\lab}. \\
    \cmidrule(lr){2-3}
    &  \multicolumn{1}{l}{\multirow{2}{*}{NSP-BERT}} & {Mapping Words}:~~~terrible; great \\
    & & {Prompt Template}:~It was {\lab}. {\sep} {\sent}  \\
    \midrule
    
    \multicolumn{1}{l}{\multirow{5}{*}{{\bf Subj}}} & & {Original Labels}:~~~~negative; positive \\
    \cmidrule(lr){2-3}
    &  \multicolumn{1}{l}{\multirow{2}{*}{PET}} & {Mapping Words}:~~~subjective; objective \\
    & & {Prompt Template}:~{\sent} This is {\lab}. \\
    \cmidrule(lr){2-3}
    &  \multicolumn{1}{l}{\multirow{2}{*}{NSP-BERT}} & {Mapping Words}:~~~subjective; objective \\
    & & {Prompt Template}:~A {\lab} comment {\sep} {\sent}  \\
    \midrule
    
    \multicolumn{1}{l}{\multirow{5}{*}{{\bf MPQA}}} & & {Original Labels}:~~~~negative; positive \\
    \cmidrule(lr){2-3}
    &  \multicolumn{1}{l}{\multirow{2}{*}{PET}} & {Mapping Words}:~~~terrible; great \\
    & & {Prompt Template}:~{\sent} It was {\lab}. \\
    \cmidrule(lr){2-3}
    &  \multicolumn{1}{l}{\multirow{2}{*}{NSP-BERT}} & {Mapping Words}:~~~negative; positive \\
    & & {Prompt Template}:~It is {\lab}. {\sep} {\sent}  \\
    \midrule
    
    \multicolumn{1}{l}{\multirow{5}{*}{{\bf Yahoo!}}} & & {Original Labels}:~~~~\makecell[l]{Society \& Culture; Science \& Mathematics; Health;\\  Education \& Reference; Computers \& Internet; Sports; \\  Business \& Finance; Entertainment \& Music; Family \\ \& Relationships; Politics \& Government} \\
    \cmidrule(lr){2-3}
    &  \multicolumn{1}{l}{\multirow{2}{*}{PET}} & {Mapping Words}:~~~\makecell[l]{Society; Science; Health; Education; Computer;\\ Sports; Business; Entertainment; Relationship; Politics} \\
    & & {Prompt Template}:~{\lab} question: {\sent}  \\
    \cmidrule(lr){2-3}
    &  \multicolumn{1}{l}{\multirow{2}{*}{NSP-BERT}} & {Mapping Words}:~~~\makecell[l]{Society; Science; Health; Education; Computer;\\ Sports; Business; Entertainment; Relationship; Politics} \\
    & & {Prompt Template}:~{\lab} question: {\sep} {\sent}  \\
    \midrule
    
    \multicolumn{1}{l}{\multirow{5}{*}{{\bf AGNews}}} & & {Original Labels}:~~~~political; sports; business; technology \\
    \cmidrule(lr){2-3}
    &  \multicolumn{1}{l}{\multirow{2}{*}{PET}} & {Mapping Words}:~~~political; sports; business; technology \\
    & & {Prompt Template}:~A {\lab} news : {\sent}  \\
    \cmidrule(lr){2-3}
    &  \multicolumn{1}{l}{\multirow{2}{*}{NSP-BERT}} & {Mapping Words}:~~~political; sports; business; technology \\
    & & {Prompt Template}:~A {\lab} news : {\sep} {\sent}  \\
    
    \bottomrule
    
    \end{tabular}
    }
    \end{center}
    \caption{The prompts used in English datasets. We only show the template with best performance. We select the most suitable prompt template for PET and NSP respectively. {\lab} is the token will be replaced by the mapping words. EFL\cite{wang2021entailment} uses the exact same prompts as NSP-BERT.}
    \label{tb:prompts_en}
    \end{table*}

%% file: tables/prompts.tex
\begin{CJK}{UTF8}{gbsn}
\begin{table*}[h]
\centering
\scriptsize
\resizebox{0.99\textwidth}{!}{
\begin{tabular}{llll}
\toprule
\textbf{Task} & \textbf{Prompt Templates} & \textbf{Mapping words of PET}  & \textbf{Mapping words of NSP-BERT}\\

\midrule

{\bf EPRSTMT} & 
\makecell[l]{
{Template 1}: \sent~{\sep} {很{\lab}.} \\
{Template 2}: \sent~{\sep} {东西很{\lab}.} \\
{Template 3}: \sent~{\sep} {这次买的东西很{\lab}}}
& \makecell[l]{好; 差} & \makecell[l]{好; 差}\\

\midrule
{\bf TNEWS} & 
\makecell[l]{
{Template 1}: \sent~\sep~\lab. \\
{Template 2}: \sent~\sep~\lab 新闻. \\
{Template 3}: \sent~\sep~这是一则 \lab 新闻.}

& \makecell[l]{
故事; 文化; 娱乐; 体育; 财经;\\  
房产; 汽车; 教育; 科技; 军事;\\
旅游; 国际; 股票; 农业; 电竞} 

& \makecell[l]{
故事; 文化; 娱乐; 体育; 财经;\\  
房产; 汽车; 教育; 科技; 军事;\\
旅游; 国际; 股票; 农业; 电竞} \\

\midrule
{\bf CSLDCP} &\makecell[l]{
{Template 1}: \sent~{\sep} {{\lab}.} \\
{Template 2}: \sent~{\sep} {{\lab} 论文.}\\
{Template 3}: \sent~{\sep} {这是一篇} {{\lab}论文.}}
& \makecell[l]{
材料; 作物; 口腔; 药学; 教育; \\ 
水利; 理经; 食品; 兽医; 体育;\\ 
核能; 力学; 园艺; 水产; 法学;  \\ 
地质; 能源; 农林; 通信; 情报...}

& \makecell[l]{
材料科学与工程; 作物学; 口腔医学; \\ 
药学; 教育学; 水利工程; 理论经济学;\\ 
食品科学与工程; 畜牧学/兽医学; \\ 
体育学; 核科学与技术; 力学; 园艺学...} \\

\midrule
{\bf IFLYTEK} & 
\makecell[l]{
{Template 1}: \sent~{\sep} {{\lab}.} \\
{Template 2}: \sent~{\sep} {{\lab} 类软件.}\\
{Template 3}: \sent~{\sep} {这是一款 {\lab} 类软件.}}
& \makecell[l]{
打车; 地图; 免费; 租车; 同城; \\
快递; 婚庆; 家政; 交通; 政务; \\
社区; 赚钱; 魔幻; 仙侠; 卡牌; \\
飞行; 射击; 休闲; 动作; 体育; \\
棋牌; 养成; 策略; 竞技; 辅助...}

& \makecell[l]{
打车; 地图导航; 免费WIFI; 租车; \\
同城服务; 快递物流; 婚庆; 家政; \\
公共交通; 政务; 社区服务; 薅羊毛;  \\
魔幻; 仙侠; 卡牌; 飞行空战; 射击游戏;\\
休闲益智; 动作类; 体育竞技...}  \\

\bottomrule
\end{tabular}
}
\vspace{-0.2cm}
\caption{The prompts used for single-sentence classification tasks in FewCLUE. {\lab} is the token will be replaced by the mapping words. The mapping words of PET need to be manually converted to equal length. Since there are two options for the prompt, {\bf prefix} and {\bf suffix}, we select the most suitable one through the development set. For dataset with a lot of labels, due to space considerations, we have omitted some of them.}
\label{tb:prompts}
\end{table*}
\end{CJK}